\documentclass[letterpaper]{article} 
\usepackage{aaai2026}  
\usepackage{times}  
\usepackage{helvet}  
\usepackage{courier}  
\usepackage[hyphens]{url}  
\usepackage{graphicx} 
\usepackage{natbib}  
\usepackage{caption} 
\usepackage{amsfonts} 
\usepackage{subcaption} 
\usepackage{booktabs} 
\usepackage{multirow} 
\usepackage[table]{xcolor} 
\usepackage{rotating} 
\usepackage{algorithm}
\usepackage{algorithmic}

\usepackage{xcolor}
\usepackage{tikz}
\usetikzlibrary{calc}
\usepackage{newfloat}
\usepackage{listings}
\DeclareCaptionStyle{ruled}{labelfont=normalfont,labelsep=colon,strut=off} 
\floatstyle{ruled}
\newfloat{listing}{tb}{lst}{}
\floatname{listing}{Listing}
\nocopyright 

\title{Trio: Learning Time-Series Forecasting with Temporal-Spatial-Sample Attention and Structural Causal Priors}
\author{
    Tao Chen\textsuperscript{\rm 1},
    Yexu Zhou\textsuperscript{\rm 1},
    Zhi Gong\textsuperscript{\rm 1},
    Hengwei He\textsuperscript{\rm 1}, 
    Hongda Li\textsuperscript{\rm 1}, 
    Zhewei Chen\textsuperscript{\rm 1}, 
    Dongjing Wang\textsuperscript{\rm 2}, 
    Xin Zhang\textsuperscript{\rm 2}, 
    Decheng Liu\textsuperscript{\rm 3}, 
    Chunlei Peng\textsuperscript{\rm 3}, 
    Zheng Chen\textsuperscript{\rm 1}, 
    Wenyue Ding\textsuperscript{\rm 1}
}
\affiliations{
    \textsuperscript{\rm 1}SUPCON Technology Co., Ltd. Hangzhou, China,\\
     \textsuperscript{\rm 2}Hangzhou Dianzi University, HangZhou, 310018, China, \\
      \textsuperscript{\rm 3}Xidian University, Xi’an 710071, Shaanxi, China.    

}

\usepackage{bibentry}

\begin{document}

\maketitle

\begin{abstract}
Multivariate time-series forecasting requires models to reason over temporal dynamics, cross-variable dependencies, and historical input-output correspondences. Recent Prior-Data Fitted Networks (PFNs) suggest that synthetic tasks can be useful for learning transferable inference behavior. However, directly transferring this paradigm to time-series forecasting remains difficult, since temporal order, dynamic lags, and recurring historical patterns are not naturally captured by ordinary tabular priors.
Motivated by this observation, we propose Trio, a sample-aware time-series forecasting architecture based on Temporal-Spatial-Sample attention. Temporal attention captures within-window dynamics, spatial attention models inter-variable dependencies, and sample attention retrieves relevant historical lookback-future pairs to guide the current prediction. Rather than claiming a fully general PFN-style forecaster, our goal is to study how historical input-output examples can be explicitly organized and reused within a forecasting model.
We further introduce a Time-Series Structural Causal Model (TS-SCM) generator to create structured synthetic forecasting tasks with dynamic lags, cross-variable interactions, noise, feedback, and distributional drift. Experiments on synthetic, industrial, and public benchmarks show that the proposed architecture improves forecasting performance. Exploratory zero-shot experiments further suggest that TS-SCM-generated tasks may provide useful structural priors, while fully general PFN-style time-series forecasting remains an open problem.
\end{abstract}


\section{Introduction}
Multivariate time-series forecasting~\cite{jung2026mambasl,zhou2024kan,hu2025timefilter}is a fundamental task in many real-world scenarios, such as energy systems, industrial monitoring, traffic prediction, and environmental analysis. Recent Transformer-based models have achieved strong performance by improving temporal modeling, patch-wise representation, and cross-variable dependency learning. However, most existing forecasting models are still trained in a dataset-specific supervised manner, where the model must be optimized separately for each target dataset. This limits their ability to learn a more general forecasting mechanism that can transfer across different time-series tasks.

\begin{figure}[t]
  \centering
  \includegraphics[width=0.9\linewidth]{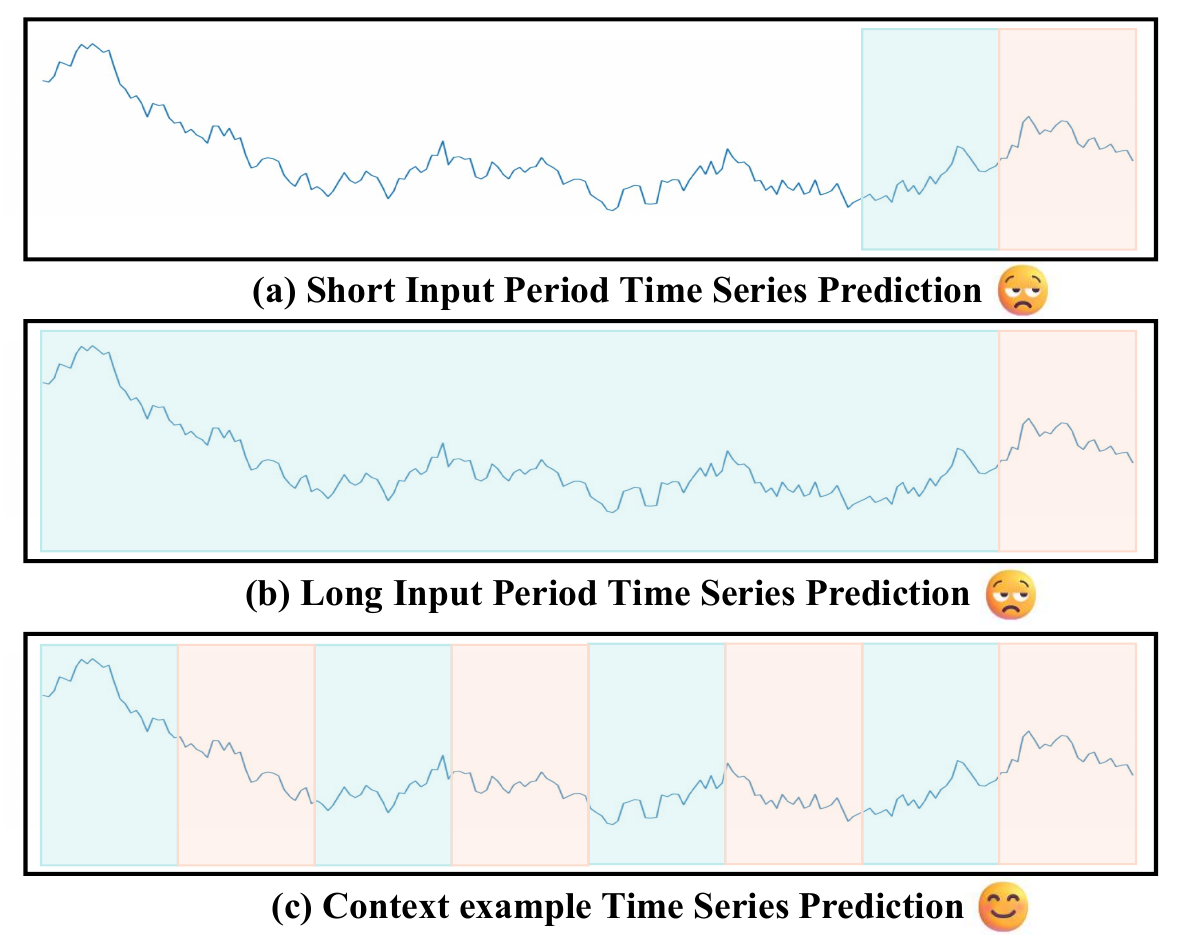}
  \caption{Three representative input paradigms: (a) short historical input; (b) long contextual input without distinguishing input–output relationships; and (c) our approach, which leverages long historical context while explicitly modeling input–output correspondences.}
  \label{fig:mov}
\end{figure}

Recently, Prior-Data Fitted Networks (PFNs), represented by TabPFN~\cite{hollmann2022tabpfn, hollmann2025accurate, grinsztajn2026tabpfn}, have shown a promising alternative paradigm in tabular learning. Instead of training from scratch on each dataset, PFN-style models learn an inference strategy from large-scale synthetic tasks generated by a prior distribution, and then transfer this learned strategy to unseen tasks. This idea has also motivated recent attempts~\cite{hoo2025tables,cai2025explore} in time-series forecasting. However, directly extending tabular PFN models to time series is insufficient, because tabular data do not explicitly contain temporal order, dynamic lagged dependencies, or historical input-output correspondences. Time-series forecasting requires models to reason jointly over temporal dynamics, cross-variable structures, and sample-level historical patterns.

Motivated by PFN-style learning, we study a more modest question: how can a forecasting model explicitly reuse historical input-output examples within a long time-series context? To this end, we propose a sample-aware forecasting architecture based on three-dimensional attention. The model decomposes multivariate time-series modeling into temporal, spatial, and sample dimensions. Temporal attention captures within-window dynamic patterns, spatial attention models dependencies among variables, and sample attention retrieves relevant historical lookback-future pairs to guide the current prediction. In this way, the model does not treat long historical context as a flat sequence, but uses historical segments as structured examples that encode how past inputs evolve into future outcomes.

In addition to the architecture, we introduce a Time-Series Structural Causal Model (TS-SCM) generator to explore whether structured synthetic tasks can serve as useful forecasting priors. Unlike simple synthetic generators that mainly produce smooth curves or weakly coupled variables, TS-SCM constructs time series through dynamic lagged dependencies, cross-variable interactions, latent factors, noise processes, error accumulation, and distributional drift. These mechanisms better reflect the structural complexity of real-world time series and provide a richer prior for learning transferable forecasting behavior.

Based on the proposed architecture and TS-SCM generator, we conduct experiments on synthetic, industrial, and public benchmark datasets. The results demonstrate that the three-dimensional attention architecture effectively improves multivariate forecasting performance. These mechanisms are intended to expose the model to structural patterns commonly observed in real-world time series, such as delayed effects, cross-variable propagation, and non-stationary transitions.
\textbf{Our main contributions are summarized as follows:}
\begin{itemize}
    \item We propose Trio, a sample-aware time-series forecasting architecture that decomposes multivariate forecasting into temporal, spatial, and sample-level attention. The model uses long histories as structured input-output examples rather than undifferentiated context.


    \item We develop TS-SCM, a forecasting-oriented synthetic task generator that simulates multivariate dynamical systems with heterogeneous node mechanisms, edge-level dynamic lags, delayed propagation, noise, feedback, and drift.
    
    \item We conduct experiments on synthetic delayed-dependency tasks, industrial datasets, and public benchmarks to validate the proposed architecture. We also include exploratory zero-shot experiments to assess whether TS-SCM-generated tasks provide useful transfer cues.
\end{itemize}

\begin{figure*}[t]
  \centering
  \includegraphics[width=0.75\linewidth]{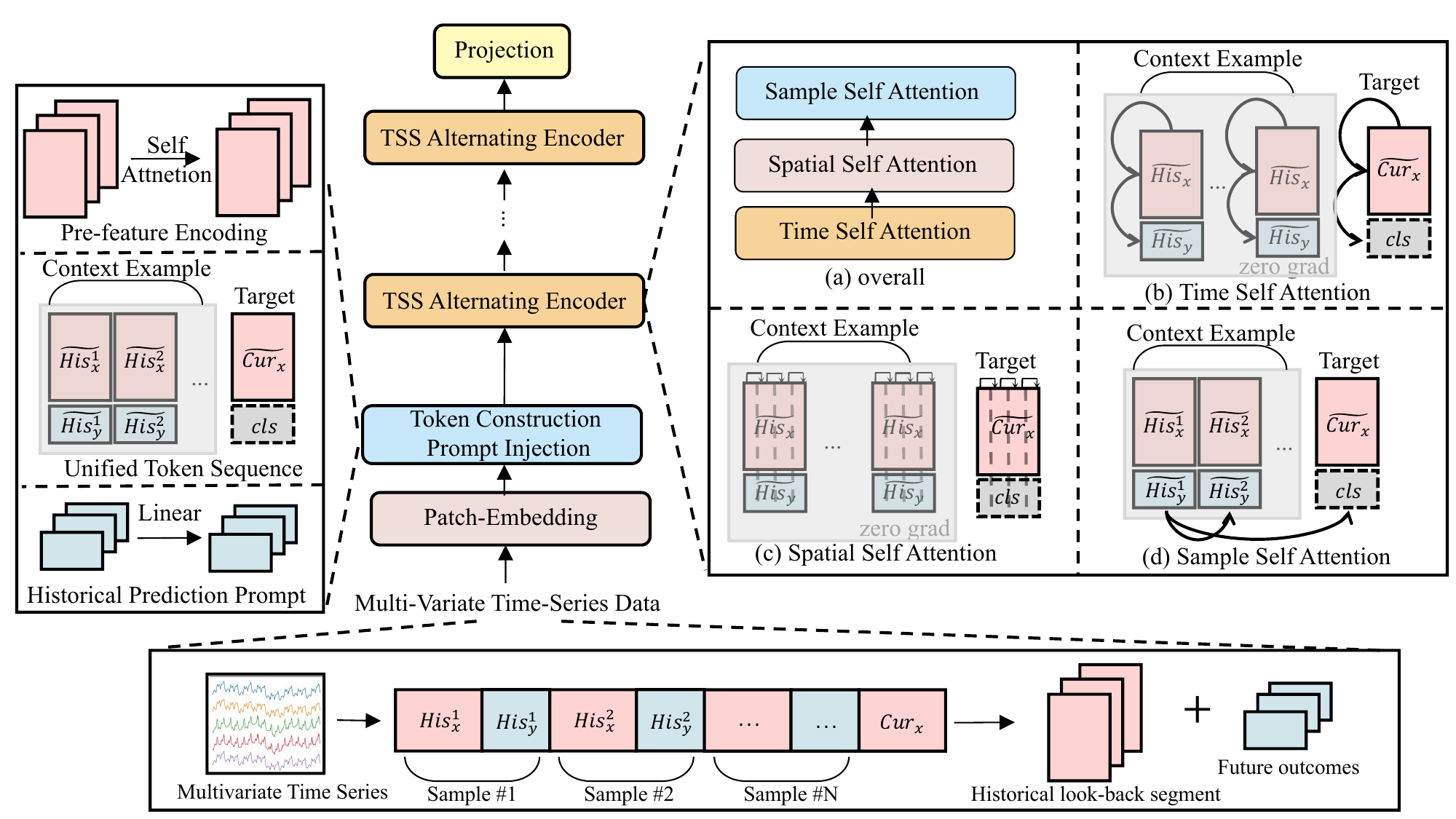}
  \caption{Overview of the Sample-Aware Transformer architecture. Our model is composed of three complementary attention mechanisms—Temporal, Spatial, and Sample-Level attention—which are applied in an alternating and hierarchical manner. Temporal attention captures dependencies across time steps, spatial attention models correlations among variables, and sample-level attention learns conditional relationships between historical inputs and future outputs.}
  \label{fig:onecol}
\end{figure*}

\section{Model Architecture}
\label{sec:model}

Given a multivariate time series
$X \in \mathbb{R}^{B \times L \times V}$,
where $B$, $L$, and $V$ denote the batch size, input length, and number of variables,
respectively, the goal is to predict a future sequence
$\hat{Y} \in \mathbb{R}^{B \times T \times V}$.
Instead of treating the entire historical context as a single continuous sequence,
we reorganize it into a set of historical input--output pairs and one current query window.

Let $L_x$ denote the length of each lookback segment and $T$ denote the prediction length.
The last $L_x$ time steps are used as the current input:
\begin{equation}
    Cur_x = X_{L-L_x+1:L} \in \mathbb{R}^{B \times L_x \times V}.
\end{equation}
The preceding context is divided into $S$ non-overlapping historical blocks, each containing
a lookback segment and its following future segment:
\begin{equation}
    (His_x^{(s)}, His_y^{(s)}), \quad s=1,\ldots,S,
\end{equation}
where
\begin{equation}
    His_x^{(s)} \in \mathbb{R}^{B \times L_x \times V}, \quad
    His_y^{(s)} \in \mathbb{R}^{B \times T \times V}.
\end{equation}
Here, each historical pair $(His_x^{(s)}, His_y^{(s)})$ provides an explicit
input--output example that describes how a previous lookback pattern evolves into its future outcome.
The model uses these historical examples as conditional evidence for predicting the future of $Cur_x$.

\paragraph{Strict temporal ordering.}
To avoid information leakage, all historical lookback-future pairs are strictly placed before the current query window. Let $L_x$ be the lookback length, $T$ be the prediction length, and $S$ be the number of historical pairs. Given an input context of length
\[
L = S(L_x+T) + L_x,
\]
we define the $s$-th historical pair as
\begin{equation}
His_x^{(s)} = X_{a_s:a_s+L_x-1}, \quad
His_y^{(s)} = X_{a_s+L_x:a_s+L_x+T-1},
\end{equation}
where
\begin{equation}
a_s = (s-1)(L_x+T)+1, \quad s=1,\ldots,S.
\end{equation}
The current input window is defined as
\begin{equation}
Cur_x = X_{S(L_x+T)+1:S(L_x+T)+L_x}.
\end{equation}
Therefore, for every historical pair,
\[
\max \mathrm{index}(His_y^{(s)}) < \min \mathrm{index}(Cur_x),
\]
which ensures that historical future prompts are past observations relative to the current window and never contain the target future of $Cur_x$.

\subsection{Patch Tokenization and Historical Prompt Construction}
\label{sec:token_construction}

We first convert both the current and historical sequences into patch-level representations.
For each variable, a patch embedding layer maps the lookback segment into $P_x$ input tokens:
\begin{equation}
    E_x = \mathrm{PatchEmbed}_x(\cdot) \in \mathbb{R}^{P_x \times D},
\end{equation}
where $D$ is the hidden dimension.
The current input is embedded as
\begin{equation}
    Cur = \mathrm{PatchEmbed}_x(Cur_x)
    \in \mathbb{R}^{B \times V \times P_x \times D}.
\end{equation}
Similarly, the historical lookback segments are embedded as
\begin{equation}
    His_x =
    \mathrm{PatchEmbed}_x(\{His_x^{(s)}\}_{s=1}^{S})
    \in \mathbb{R}^{B \times V \times S \times P_x \times D}.
\end{equation}

Different from simply compressing the historical future into a single vector,
we preserve its patch-wise structure. Specifically, the historical future segment
$His_y^{(s)}$ is encoded by a separate future patch embedding layer:
\begin{equation}
    His_y =
    \mathrm{PatchEmbed}_y(\{His_y^{(s)}\}_{s=1}^{S})
    \in \mathbb{R}^{B \times V \times S \times P_y \times D},
\end{equation}
where $P_y$ denotes the number of future tokens.
These future tokens act as \emph{historical prediction prompts}, providing patch-wise supervision signals
that indicate how each historical lookback window evolves into its future.

Before constructing the final token sequence, we apply lightweight self-attention-based
pre-encoders to improve intra-window representations:
\begin{equation}
    \widetilde{Cur} = \mathrm{PreEnc}_x(Cur),
\end{equation}
\begin{equation}
    \widetilde{His}_x = \mathrm{PreEnc}_x(His_x), \quad
    \widetilde{His}_y = \mathrm{PreEnc}_y(His_y).
\end{equation}
The two pre-encoders operate along the patch dimension and are used to enhance local
lookback and future-prompt representations before cross-window interaction.

For each historical window, we concatenate the encoded lookback tokens and future-prompt tokens:
\begin{equation}
    His_{\mathrm{tok}}^{(s)}
    =
    [\widetilde{His}_x^{(s)}, \widetilde{His}_y^{(s)}]
    \in \mathbb{R}^{B \times V \times (P_x+P_y) \times D}.
\end{equation}

For the current window, since the future is unknown, we append $P_y$ learnable query tokens:
\begin{equation}
    Q_y \in \mathbb{R}^{1 \times V \times P_y \times D}.
\end{equation}
The current tokens are then constructed as
\begin{equation}
    Cur_{\mathrm{tok}}
    =
    [\widetilde{Cur}, Q_y]
    \in \mathbb{R}^{B \times V \times (P_x+P_y) \times D}.
\end{equation}
The learnable query tokens play the role of future placeholders. During sample-level retrieval,
they query historical future-prompt tokens and absorb useful input--output mappings from previous windows.

Finally, all historical tokens and the current tokens are stacked along the sample/window dimension:
\begin{equation}
    Z^{(0)}
    =
    [His_{\mathrm{tok}}^{(1)}, \ldots, His_{\mathrm{tok}}^{(S)}, Cur_{\mathrm{tok}}]
    \in \mathbb{R}^{B \times V \times (S+1) \times (P_x+P_y) \times D}.
\end{equation}
For notation simplicity, we denote the total number of windows as
$\bar{S}=S+1$ and the token length within each window as $P=P_x+P_y$.

\subsection{Temporal-Spatial-Sample Alternating Encoder}
\label{sec:tss_encoder}

The constructed tensor $Z^{(0)} \in \mathbb{R}^{B \times V \times \bar{S} \times P \times D}$
contains three types of dependencies:
temporal dependencies along the patch dimension,
spatial dependencies among variables,
and sample-level dependencies among historical windows and the current query window.
Directly applying full attention over all axes is computationally expensive and introduces
uncontrolled interactions. Therefore, we design a Temporal-Spatial-Sample alternating encoder,
which decomposes the modeling process into axis-wise attention operations.

Each encoder block contains three modules: spatial attention, temporal attention, and sample attention.
The overall update can be written as
\begin{equation}
    Z^{(k)}
    =
    F_{\mathrm{sample}}^{(k)}
    \circ
    F_{\mathrm{temp}}^{(k)}
    \circ
    F_{\mathrm{spat}}^{(k)}
    (Z^{(k-1)}),
    \quad k=1,\ldots,K.
\end{equation}

\paragraph{Spatial Attention.}
The spatial module models dependencies among variables at each window and patch position.
Given $Z \in \mathbb{R}^{B \times V \times \bar{S} \times P \times D}$,
we reshape it as
\begin{equation}
    Z_{\mathrm{spat}}
    =
    \mathrm{reshape}(Z)
    \in \mathbb{R}^{(B\bar{S}P) \times V \times D}.
\end{equation}
Non-causal self-attention is then applied along the variable dimension:
\begin{equation}
    \widehat{Z}_{\mathrm{spat}}
    =
    \mathrm{Attn}_{V}(Z_{\mathrm{spat}}).
\end{equation}
The output is reshaped back to
$\mathbb{R}^{B \times V \times \bar{S} \times P \times D}$.
This operation allows each variable to dynamically aggregate information from other variables
at the same sample and patch position.

For dataset-specific supervised training with long contexts, we optionally apply a zero-gradient strategy to reduce memory cost. 
Specifically, spatial attention for historical windows is first computed under \texttt{no\_grad}, while the current window is recomputed with gradient tracking and written back to the output. 
In this setting, historical windows are treated as contextual evidence, and parameter updates are mainly driven by the current prediction loss. 
This option is not required in all training settings and is disabled or adjusted when full-gradient optimization is needed.

\paragraph{Temporal Attention.}
The temporal module captures within-window patch dependencies for each variable and sample.
The tensor is reshaped as
\begin{equation}
    Z_{\mathrm{temp}}
    =
    \mathrm{reshape}(Z)
    \in \mathbb{R}^{(BV\bar{S}) \times P \times D}.
\end{equation}
We apply causal self-attention along the patch dimension:
\begin{equation}
    \widehat{Z}_{\mathrm{temp}}
    =
    \mathrm{Attn}^{\mathrm{causal}}_{P}(Z_{\mathrm{temp}}).
\end{equation}
The causal mask prevents later tokens from being accessed by earlier tokens.
When the optional zero-gradient strategy is enabled for long-context supervised training, 
historical windows are processed under \texttt{no\_grad}, whereas the current window is recomputed with gradients and replaces its corresponding position in the output tensor. 
Otherwise, the temporal module can be optimized in a standard full-gradient 

\paragraph{Sample Attention.}
The sample module is designed to retrieve useful historical input--output mappings for the current prediction.
Unlike conventional sample attention that only uses a single CLS token, our implementation performs
patch-wise sample retrieval over the future-query tokens.

Let the last $P_y$ tokens of each window be denoted as
\begin{equation}
    R = Z[:, :, :, -P_y:, :]
    \in \mathbb{R}^{B \times V \times \bar{S} \times P_y \times D}.
\end{equation}
For historical windows, we take
\begin{equation}
    R_{\mathrm{hist}}
    =
    R[:, :, 1:S, :, :]
    \in \mathbb{R}^{B \times V \times S \times P_y \times D}.
\end{equation}
The historical future-prompt tokens are then rearranged as
\begin{equation}
    R_{\mathrm{hist}}
    \rightarrow
    \mathbb{R}^{(BVP_y) \times S \times D},
\end{equation}
so that sample self-attention is independently performed for each variable and each future-patch position:
\begin{equation}
    \widehat{R}_{\mathrm{hist}}
    =
    \mathrm{Attn}_{S}(R_{\mathrm{hist}}).
\end{equation}
This step aggregates historical windows and produces a memory bank of historical future-prompt representations.

Next, the current future-query tokens are extracted:
\begin{equation}
    R_{\mathrm{cur}}
    =
    R[:, :, -1, :, :]
    \in \mathbb{R}^{B \times V \times P_y \times D}.
\end{equation}
After reshaping it into
$\mathbb{R}^{(BVP_y) \times 1 \times D}$,
we use it as the query and use the aggregated historical tokens as keys and values:
\begin{equation}
    \widehat{R}_{\mathrm{cur}}
    =
    \mathrm{CrossAttn}
    \left(
    R_{\mathrm{cur}},
    \widehat{R}_{\mathrm{hist}},
    \widehat{R}_{\mathrm{hist}}
    \right).
\end{equation}
Only the current future-query tokens are updated and written back to $Z$:
\begin{equation}
    Z[:, :, -1, -P_y:, :]
    \leftarrow
    \widehat{R}_{\mathrm{cur}}.
\end{equation}
This realizes a history-first aggregation and current-only retrieval mechanism.
It avoids bidirectional mixing between historical and current windows, while allowing the current prediction
tokens to selectively retrieve relevant historical input--output patterns.

\subsection{Prediction Head}
\label{sec:prediction_head}

After the TSS alternating encoder, we extract all tokens from the current window:
\begin{equation}
    H
    =
    Z^{(K)}[:, :, -1, :, :]
    \in \mathbb{R}^{B \times V \times (P_x+P_y) \times D}.
\end{equation}
Instead of using only a single CLS representation, we flatten all current-window tokens for each variable:
\begin{equation}
    h
    =
    \mathrm{Flatten}(H)
    \in \mathbb{R}^{B \times V \times ((P_x+P_y)D)}.
\end{equation}
A shared linear head maps the representation of each variable to the prediction horizon:
\begin{equation}
    \hat{Y}
    =
    \mathrm{Head}(h)
    \in \mathbb{R}^{B \times T \times V}.
\end{equation}

When probabilistic forecasting is used, the head outputs $Q$ quantiles:
\begin{equation}
    \hat{Y}
    =
    \mathrm{Head}(h)
    \in \mathbb{R}^{B \times T \times V \times Q}.
\end{equation}
For point forecasting, $Q=1$ or the median quantile is used as the final prediction.

Following common practice in non-stationary time-series forecasting, we normalize each input sequence
before tokenization and apply inverse normalization to the prediction:
\begin{equation}
    \bar{X} = \frac{X-\mu}{\sigma},
    \quad
    \hat{Y} = \sigma \cdot \hat{\bar{Y}} + \mu,
\end{equation}
where $\mu$ and $\sigma$ are computed from the input lookback sequence.

For deterministic forecasting, the model is optimized by minimizing the mean squared error:
\begin{equation}
    \mathcal{L}
    =
    \frac{1}{BTV}
    \sum_{b=1}^{B}
    \sum_{t=1}^{T}
    \sum_{v=1}^{V}
    \left(
    \hat{Y}_{b,t,v} - Y_{b,t,v}
    \right)^2.
\end{equation}
For quantile forecasting, this objective can be replaced by the standard pinball loss over all quantile levels.

\section{Time-Series Structural Causal Model Generator}
\label{sec:tsscm}

To provide structured synthetic tasks for PFN-style time-series forecasting, 
we introduce a Time-Series Structural Causal Model (TS-SCM) generator.
Different from GP-based synthetic priors that mainly define a distribution over temporal functions, 
TS-SCM defines a distribution over multivariate causal dynamical systems. 
Each generated task is specified by a directed graph, heterogeneous node mechanisms, edge-level lag functions, noise processes, and optional mechanism drift.

\paragraph{Graph and node sampling.}
TS-SCM first samples a set of root variables and affected variables.
Root variables act as exogenous driving processes and can be generated from periodic functions, Fourier components, trends, Gaussian-process-like priors, square-wave patterns, or chaotic systems.
Affected variables are generated by node-wise mechanisms that receive delayed inputs from their parent variables.
The supported node mechanisms include linear functions, bounded nonlinear unary functions, kernel RBF functions, and other nonlinear transformations.
For an affected node $x_i$, the basic update can be written as
\begin{equation}
    x_i(t) =
    \alpha_i x_i(t-1)
    +
    \beta_i f_i\left(
        m_{1i}(t), \ldots, m_{|Pa(i)|i}(t)
    \right)
    +
    \epsilon_i(t),
\end{equation}
where $Pa(i)$ denotes the parent set of node $i$, 
$f_i(\cdot)$ is the node-specific mechanism,
$m_{ji}(t)$ is the message delivered from parent $x_j$ to child $x_i$,
$\alpha_i$ controls state persistence,
$\beta_i$ controls input strength,
and $\epsilon_i(t)$ denotes noise.

\paragraph{Edge-level causal messages.}
Each directed edge $e_{ji}: x_j \rightarrow x_i$ specifies how information from the parent variable is transformed and delivered to the child variable.
The message on an edge is defined by an edge weight, an optional nonlinear transform, a lag function, a reduce policy, and a fill policy.
Abstractly, the delivered message can be written as
\begin{equation}
    m_{ji}(t)
    =
    \rho_{ji}
    \left(
    \left\{
    w_{ji}(t) \cdot 
    \psi_{ji}\left(x_j(t')\right)
    \; \middle| \;
    t' + \tau_{ji}(t') = t
    \right\}
    \right),
\end{equation}
where $\psi_{ji}(\cdot)$ is the edge transform,
$w_{ji}(t)$ is the possibly drifted edge weight,
$\tau_{ji}(\cdot)$ is the lag function,
and $\rho_{ji}(\cdot)$ aggregates multiple events arriving at the same time.
The reduce policy can be sum, mean, max, first, or last.
When no event arrives at a time step, TS-SCM supports different fill policies, including zero filling, forward filling, and linear interpolation.

\begin{table*}[t]
\centering
\caption{Comparison and ablation results on two synthetic datasets.
Lower values indicate better performance.}
\label{tab:synthetic_ablation}
\begin{tabular}{cc}
\begin{minipage}{0.48\linewidth}
\centering
\subcaption{Synthetic Dataset~\#1, where $Y$ is an $M$-step delayed version of $X$, with $M \sim \mathcal{U}(20, 60)$.}
\begin{tabular}{lcc}
\toprule
Method & MSE $\downarrow$ & MAE $\downarrow$ \\
\midrule
\multicolumn{3}{l}{\emph{Standard baselines}} \\
TimesNet    & 0.826 & 0.725 \\
Crossformer  & 0.980 & 0.503 \\
TimeXer      & 0.736 & 0.684 \\
\midrule
\multicolumn{3}{l}{\emph{Ablation on sample attention}} \\
Base Model (w/o Sample Attn.) & 0.706 & 0.670 \\
\textbf{Base Model + Sample Attn.} 
                             & \textbf{0.298} & \textbf{0.431} \\
\bottomrule
\end{tabular}
\end{minipage}
&
\begin{minipage}{0.48\linewidth}
\centering
\subcaption{Synthetic Dataset~\#2, where $Y$ is an $M$-step delayed version of one randomly selected variable from $\{X_1,\dots,X_5\}$, with $M \sim \mathcal{U}(20, 30)$.}
\begin{tabular}{lcc}
\toprule
Method & MSE $\downarrow$ & MAE $\downarrow$ \\
\midrule
\multicolumn{3}{l}{\emph{Standard baselines}} \\
TimesNet     & 0.981 & 0.789 \\
Crossformer  & 0.795 & 0.710 \\
TimeXer      & 0.956 & 0.779 \\
\midrule
\multicolumn{3}{l}{\emph{Ablation on sample attention}} \\
Base Model (w/o Sample Attn.) & 1.011 & 0.802 \\
\textbf{Base Model + Sample Attn.}
                             & \textbf{0.644} & \textbf{0.638} \\
\bottomrule
\end{tabular}
\end{minipage}
\end{tabular}~\label{tab:synthetic_ablation}
\end{table*}

\begin{table*}[t]
\centering
\caption{Comparison results on benchmark datasets under different forecasting horizons.
Metric is Mean Squared Error (MSE); lower is better.}
\label{tab:comparison_mse}

\small
\setlength{\tabcolsep}{3.5pt}
\renewcommand{\arraystretch}{1.08}

\resizebox{\textwidth}{!}{
\begin{tabular}{llcccccccccc}
\toprule
Dataset & Horizon
& \textbf{Trio}
& TimeMixer++
& Patch-wise
& TIMEKAN
& MOIRAI
& TimesFM
& TimeMixer
& iTransformer
& PatchTST
& DLinear \\
\midrule

\multirow{5}{*}{ETTm1}
& 96  & \textbf{0.289} & 0.310 & 0.326 & 0.322 & 0.353 & 0.345 & 0.320 & 0.334 & 0.352 & 0.346 \\
& 192 & \textbf{0.339} & 0.348 & 0.374 & 0.357 & 0.376 & 0.374 & 0.361 & 0.390 & 0.374 & 0.382 \\
& 336 & 0.377 & \textbf{0.376} & 0.410 & 0.382 & 0.399 & 0.397 & 0.390 & 0.426 & 0.421 & 0.415 \\
& 720 & \textbf{0.428} & 0.440 & 0.472 & 0.445 & 0.432 & 0.436 & 0.454 & 0.491 & 0.462 & 0.473 \\
& Avg & \textbf{0.358} & 0.369 & 0.396 & 0.377 & 0.390 & 0.388 & 0.381 & 0.410 & 0.402 & 0.404 \\
\midrule

\multirow{5}{*}{ETTm2}
& 96  & \textbf{0.156} & 0.170 & 0.175 & 0.174 & 0.189 & 0.263 & 0.175 & 0.180 & 0.183 & 0.193 \\
& 192 & \textbf{0.219} & 0.229 & 0.242 & 0.239 & 0.247 & 0.309 & 0.237 & 0.250 & 0.255 & 0.284 \\
& 336 & \textbf{0.285} & 0.303 & 0.304 & 0.301 & 0.295 & 0.349 & 0.298 & 0.311 & 0.309 & 0.382 \\
& 720 & \textbf{0.360} & 0.373 & 0.401 & 0.395 & 0.372 & 0.415 & 0.391 & 0.412 & 0.412 & 0.558 \\
& Avg & \textbf{0.255} & 0.269 & 0.281 & 0.277 & 0.276 & 0.334 & 0.275 & 0.288 & 0.290 & 0.354 \\
\midrule

\multirow{5}{*}{Electricity}
& 96  & \textbf{0.133} & 0.135 & 0.146 & 0.174 & 0.152 & -- & 0.153 & 0.148 & 0.190 & 0.210 \\
& 192 & 0.156 & \textbf{0.147} & 0.161 & 0.182 & 0.171 & -- & 0.166 & 0.162 & 0.199 & 0.210 \\
& 336 & 0.172 & \textbf{0.164} & 0.174 & 0.197 & 0.192 & -- & 0.185 & 0.178 & 0.217 & 0.223 \\
& 720 & \textbf{0.195} & 0.212 & 0.208 & 0.236 & 0.236 & -- & 0.225 & 0.225 & 0.258 & 0.258 \\
& Avg & \textbf{0.164} & 0.165 & 0.172 & 0.197 & 0.188 & -- & 0.182 & 0.178 & 0.216 & 0.225 \\
\midrule

\multirow{5}{*}{Weather}
& 96  & \textbf{0.149} & 0.155 & 0.167 & 0.162 & 0.177 & -- & 0.163 & 0.174 & 0.186 & 0.195 \\
& 192 & 0.212 & \textbf{0.201} & 0.219 & 0.207 & 0.219 & -- & 0.208 & 0.221 & 0.234 & 0.237 \\
& 336 & 0.270 & \textbf{0.237} & 0.274 & 0.263 & 0.277 & -- & 0.251 & 0.278 & 0.284 & 0.282 \\
& 720 & 0.364 & \textbf{0.312} & 0.353 & 0.338 & 0.365 & -- & 0.339 & 0.358 & 0.356 & 0.345 \\
& Avg & 0.249 & \textbf{0.226} & 0.253 & 0.243 & 0.260 & -- & 0.240 & 0.258 & 0.265 & 0.265 \\
\bottomrule
\end{tabular}
}
\end{table*}

\paragraph{Dynamic lag scheduling.}
A central design of TS-SCM is that lags are defined at the edge level rather than by a global delay matrix.
For each edge, the lag can be fixed, state-dependent, history-dependent, sampled from a value-dependent interval, or defined by a user-specified expression:
\begin{equation}
    \tau_{ji}(t)
    =
    g_{ji}\left(x_j(t), t, \mathcal{H}_j(t)\right),
\end{equation}
where $\mathcal{H}_j(t)$ denotes the history of the parent variable.
This formulation allows different causal effects to propagate with different and time-varying delays.
For example, an edge can have a larger delay when the parent value is large, or switch between different delay ranges according to the current state.

\paragraph{Push-based event calendar.}
Instead of letting the child node pull a parent value from a fixed past index, TS-SCM uses a push-based event calendar.
When a parent value $x_j(t)$ is generated, each outgoing edge computes a future arrival trajectory
\begin{equation}
    s_{ji}(t) = t + \tau_{ji}(t).
\end{equation}
Whenever the continuous trajectory $s_{ji}(t)$ crosses an integer time bucket $k$, an event is scheduled to arrive at $k$.
The parent value attached to that event is linearly interpolated between consecutive parent states:
\begin{equation}
    \tilde{x}_j(k)
    =
    (1-\lambda)x_j(t-1) + \lambda x_j(t),
\end{equation}
where $\lambda$ is determined by the crossing location.
The event is then stored in the calendar of edge $e_{ji}$ and consumed by the child node at its arrival time.
This design avoids the crude rounding of continuous lags and enables smooth delayed propagation under time-varying lag functions.

\paragraph{Zero-lag and delayed recurrent dependencies.}
TS-SCM distinguishes synchronous dependencies from delayed dependencies.
Zero-lag edges are allowed only when explicitly specified, and the zero-lag subgraph is constrained to be acyclic so that nodes at the same time step have a valid computation order.
In contrast, positive-lag edges can form recurrent feedback loops, including self-loops and reverse-direction effects, because their messages are delivered to future time steps.
This separation allows TS-SCM to model both instantaneous interactions and delayed feedback while avoiding ill-defined simultaneous cycles.

\paragraph{Drift, noise, and stability control.}
To simulate nonstationary real-world systems, TS-SCM supports piecewise mechanism drift.
The drift engine can modify node parameters, global linear scales, or edge weights over time.
Noise can be Gaussian, Student-$t$, autoregressive, or mixture noise, allowing both regular disturbance and heavy-tailed perturbation.
To prevent randomly sampled recurrent systems from becoming unstable, TS-SCM applies stability-aware graph sampling.
In particular, incoming edge gains are controlled by a child-wise budget, and self-feedback edges use a stricter gain budget than cross-variable delayed edges.
This produces diverse yet learnable synthetic systems rather than arbitrary unstable trajectories.

\paragraph{Forecasting episode construction.}
Finally, each generated trajectory is converted into forecasting episodes.
Given a long trajectory, TS-SCM constructs multiple historical lookback--future pairs and a current query window:
\begin{equation}
    \mathcal{E}
    =
    \left\{
    (His_x^{(1)}, His_y^{(1)}),
    \ldots,
    (His_x^{(S)}, His_y^{(S)}),
    Cur_x
    \right\}.
\end{equation}
This episode format is directly aligned with our three-dimensional attention architecture:
historical lookback--future pairs provide sample-level evidence,
while the current query window retrieves relevant input--output mappings for prediction.
Therefore, TS-SCM is not only a trajectory generator, but also a forecasting-oriented task generator.

\section{Experiments}

\subsection{Experimental Settings}

We evaluate Trio on three groups of datasets: synthetic delayed-dependency datasets, and public forecasting benchmarks. The synthetic datasets are used to isolate delayed temporal dependencies and variable-selection uncertainty. The public benchmarks include ETTm1, ETTm2, Weather, and Electricity.

For public benchmarks, we follow the standard~\cite{wang2024tssurvey} train/validation/test splits used in prior long-term forecasting studies. All methods are evaluated under the same prediction horizons. For our model, the input context is organized into historical lookback-future pairs and one current query window according to the indexing rule described in Sec.~\ref{sec:model}. For baselines, we use the context lengths and hyperparameters reported in their original papers whenever available, and otherwise tune them on the validation set.

All experiments are implemented in PyTorch and trained with the Adam optimizer. The initial learning rate is set to $10^{-4}$. We use the mean squared error loss for deterministic forecasting and apply early stopping based on the validation loss. Unless otherwise specified, results are reported using MSE and MAE, where lower values indicate better performance. For experiments involving randomness, we use fixed random seeds and report the best validation-selected checkpoint on the test set.

\subsection{Comparison Methods}

We compare Trio with representative forecasting baselines, including TimeMixer++~\cite{wang2025timemixer++}, Patch-wise Structural~\cite{kudrat2025patchwise}, TIMEKAN~\cite{huang2025timekan}, MOIRAI~\cite{woo2024moirai}, TimesFM~\cite{10.5555/3692070.3692474}, TimeMixer~\cite{wang2024timemixer}, iTransformer~\cite{liu2024itransformer}, PatchTST~\cite{nie2022time}, DLinear~\cite{zeng2023transformers}. 
These baselines cover Transformer-based, patch-based, decomposition-based, MLP-based, and foundation-style forecasting methods. 
All results are reported under multiple prediction horizons for a comprehensive comparison.

\subsection{Results on Synthetic Scenarios}

We first validate the proposed sample-level modeling strategy on two synthetic datasets.
In Synthetic Dataset~\#1, the target series $Y$ is generated as an $M$-step delayed version of the input series $X$, where the delay is sampled from $M \sim \mathcal{U}(20,60)$.
This setting is used to test whether a model can capture variable temporal delays.
In Synthetic Dataset~\#2, $Y$ is generated as an $M$-step delayed version of one randomly selected variable from the multivariate input $\{X_1,\ldots,X_5\}$, where $M \sim \mathcal{U}(20,30)$.
This setting is more challenging because the model must identify not only the delay length but also the relevant source variable.

As shown in Table~\ref{tab:synthetic_ablation}, standard forecasting baselines show limited performance under these delayed-dependency settings.
In contrast, adding Sample Attention consistently improves the base model.
On Synthetic Dataset~\#1, the MSE decreases from 0.706 to 0.298.
On Synthetic Dataset~\#2, the MSE decreases from 1.011 to 0.644.
The improvement is more pronounced in the second setting, suggesting that sample-level retrieval is particularly useful when the model needs to infer both temporal delay and variable relevance.
These results support our motivation that historical lookback--future pairs provide useful evidence for current prediction.

\subsection{Results on Public Benchmarks}

As shown in Table~\ref{tab:comparison_mse}, Trio achieves the best average MSE on three out of four datasets, including ETTm1, ETTm2, and Electricity. 
Specifically, Trio obtains average MSE scores of 0.358, 0.255, and 0.164 on ETTm1, ETTm2, and Electricity, respectively. 
On Weather, Trio achieves an average MSE of 0.249, which is not the best overall result; TimeMixer++ obtains the lowest average MSE of 0.226 on this dataset. 
Nevertheless, Trio remains competitive on Weather and achieves the best result at the short prediction horizon $H=96$. 
These results suggest that Trio can effectively exploit historical input--output correspondences on most benchmarks, while its advantage may vary across datasets with different temporal characteristics.

\begin{table}[h]
\centering
\caption{Comparison of ICTSP and Trio on ETTm1 and ETTm2 datasets with different context lengths.}
\label{tab:comparison_ICTSP}

\scalebox{0.7}{
\begin{tabular}{c|c|c|c|c|c|c|c|c}

\hline
Dataset & \multicolumn{4}{c|}{ETTm1} & \multicolumn{4}{c}{ETTm2} \\
\hline

\multirow{2}{*}{Context Length} 
& \multicolumn{2}{c|}{ICTSP} 
& \multicolumn{2}{c|}{Trio} 
& \multicolumn{2}{c|}{ICTSP} 
& \multicolumn{2}{c}{Trio} \\
\cline{2-9}

& Mean & Best & Mean & Best & Mean & Best & Mean & Best \\
\hline 

2000 
& 0.381 
& \multirow{3}{*}{0.370} 
& \textbf{0.358} 
& \multirow{3}{*}{\textbf{0.358}} 
& 0.265 
& \multirow{3}{*}{0.262} 
& \textbf{0.255} 
& \multirow{3}{*}{\textbf{0.255}} \\

\cline{1-2} \cline{4-4} \cline{6-6} \cline{8-8}

3000 
& 0.371 
&  
& \textbf{0.359} 
&  
& 0.262 
&  
& \textbf{0.256} 
&  \\

\cline{1-2} \cline{4-4} \cline{6-6} \cline{8-8}

4000 
& 0.370 
&  
& \textbf{0.365} 
&  
& \textbf{0.266} 
&  
& \textbf{0.266} 
&  \\
\hline

\end{tabular}
}
\end{table}

We further compare our model with ICTSP~\cite{lu2024context} under different context lengths in Table~\ref{tab:comparison_ICTSP}.
ICTSP also attempts to use long historical context for forecasting, but it does not explicitly organize historical segments as input--output examples.
In contrast, our model uses historical lookback--future pairs and sample-level retrieval to model the relationship between past observations and future outcomes.
As a result, our method achieves better best performance on both ETTm1 and ETTm2.
This comparison supports the advantage of explicitly modeling historical input--output mappings rather than simply increasing the context length.

\subsection{Ablation Study}

\begin{table}[t]
\centering
\caption{Ablation results on ETTm1 and ETTm2 datasets. We compare the base model performance with and without Sample Attention across various context and prediction lengths. Metric is Mean Squared Error (MSE) and Mean Absolute Error (MAE); lower is better.}
\label{tab:ablation_sample_attn_scaled}

\scriptsize

\scalebox{1.0}{                     
\begin{tabular}{@{}lccccccc@{}}
\toprule
\multirow{2}{*}{Dataset} & \multirow{2}{*}{Context / Horizon} & \multicolumn{2}{c}{Base Model} & & \multicolumn{2}{c}{\textbf{Base + Sample Attn}} \\
\cmidrule{3-4} \cmidrule{6-7}
& & MSE $\downarrow$ & MAE $\downarrow$ & & MSE $\downarrow$ & MAE $\downarrow$ \\
\midrule

\multirow{4}{*}{ETTm1} 
& 1056 / 96  & 0.323 & 0.379 & & \textbf{0.306} & \textbf{0.361} \\
& 1536 / 192 & 0.374 & 0.414 & & \textbf{0.347} & \textbf{0.394} \\
& 2256 / 336 & 0.444 & 0.450 & & \textbf{0.369} & \textbf{0.409} \\
& 4176 / 720 & 0.512 & 0.521 & & \textbf{0.424} & \textbf{0.446} \\
\midrule

\multirow{4}{*}{ETTm2} 
& 1056 / 96  & 0.191 & 0.284 & & \textbf{0.175} & \textbf{0.270} \\
& 1536 / 192 & 0.294 & 0.352 & & \textbf{0.230} & \textbf{0.310} \\
& 2256 / 336 & 0.345 & 0.392 & & \textbf{0.291} & \textbf{0.349} \\
& 4176 / 720 & 0.394 & 0.434 & & \textbf{0.383} & \textbf{0.434} \\
\bottomrule
\end{tabular}
}
\end{table}

\paragraph{Definition of the Base Model.}
In all ablation studies, the Base Model refers to the same backbone as Trio except that the Sample Attention module is removed. Specifically, it keeps the same patch embedding layers, pre-encoders, temporal attention, spatial attention, prediction head, normalization scheme, model depth, hidden dimension, and training protocol. The only difference is that the current future-query tokens do not retrieve historical future prompts through the sample dimension. Therefore, the comparison between the Base Model and Base Model + Sample Attention isolates the contribution of sample-level historical input--output retrieval.
\paragraph{Effectiveness of Sample Attention.}
Table~\ref{tab:ablation_sample_attn_scaled} evaluates the contribution of Sample Attention on ETTm1 and ETTm2.
Across all tested horizons, adding Sample Attention consistently improves the base model.
For example, on ETTm1 with horizon 720, the MSE decreases from 0.512 to 0.424.
On ETTm2 with horizon 192, the MSE decreases from 0.294 to 0.230.
These improvements demonstrate that retrieving relevant historical lookback--future pairs helps the model capture long-range and delayed temporal dependencies.

\paragraph{Scalability to Long Contexts.}

\begin{table}[t]
\centering
\caption{Average performance across all prediction horizons under different context lengths with the default input length of 96. Lower values indicate better performance.}
\label{tab:long_context_avg}

\scalebox{0.55}{ 
\setlength{\tabcolsep}{4pt}
\renewcommand{\arraystretch}{1.08}

\begin{tabular}{c|cc|cc|cc|cc|cc|cc}
\toprule
\multirow{2}{*}{Context Length ($L$)}
& \multicolumn{2}{c|}{\textbf{ETTm1}}
& \multicolumn{2}{c|}{\textbf{ETTm2}}
& \multicolumn{2}{c|}{\textbf{Weather}}
& \multicolumn{2}{c|}{\textbf{ECL}}
& \multicolumn{2}{c|}{\textbf{ETTh1}}
& \multicolumn{2}{c}{\textbf{ETTh2}} \\
\cmidrule{2-13}
& MSE & MAE & MSE & MAE & MSE & MAE & MSE & MAE & MSE & MAE & MSE & MAE \\
\midrule
96
& 0.387 & 0.391
& 0.281 & 0.320
& 0.252 & 0.274
& 0.167 & 0.259
& 0.450 & 0.433
& 0.376 & 0.397 \\

1000
& \textbf{0.356} & \textbf{0.378}
& 0.263 & 0.318
& 0.249 & \textbf{0.273}
& \textbf{0.164} & \textbf{0.257}
& \textbf{0.440} & \textbf{0.434}
& \textbf{0.353} & \textbf{0.389} \\

2000
& 0.358 & 0.385
& \textbf{0.255} & 0.318
& 0.248 & 0.275
& 0.167 & 0.264
& 0.477 & 0.452
& 0.366 & 0.404 \\

3000
& 0.359 & 0.385
& 0.256 & \textbf{0.315}
& 0.247 & 0.276
& 0.168 & 0.264
& -- & --
& -- & -- \\

4000
& 0.365 & 0.393
& 0.266 & 0.320
& \textbf{0.242} & 0.278
& 0.172 & 0.268
& --& --
& -- & -- \\
\bottomrule
\end{tabular}
}
\end{table}

Tables~\ref{tab:long_context_avg} and~\ref{tab:long_context_avg_336} report the average performance across all prediction horizons under different context lengths, with default input lengths of 96 and 336, respectively. 
Increasing the context length generally improves forecasting performance in both settings, showing the benefit of using longer historical contexts. 
However, the best context length varies across datasets and input settings. 
For the default input length of 96, moderate context lengths such as $L=1000$ or $L=2000$ already provide strong performance, while further increasing $L$ does not consistently improve the results. 
For the default input length of 336, ETTm1 and ETTm2 obtain the best average performance around $L=3000$, but performance drops when $L$ is further increased. 
This may be because longer input windows reduce the number of available training samples and may also introduce redundant or noisy historical information. 
Therefore, moderate long-context modeling is useful, but blindly increasing the context length is not always beneficial. 
Full horizon-wise results are provided in Appendix.

\begin{table}[t]
\centering
\caption{Average performance across all prediction horizons under different context lengths with the default input length of 336. Lower values indicate better performance.}
\label{tab:long_context_avg_336}

\small
\setlength{\tabcolsep}{5pt}
\renewcommand{\arraystretch}{1.08}
\scalebox{0.7}{ 

\begin{tabular}{c|cc|cc|cc|cc}
\toprule
\multirow{2}{*}{Context Length ($L$)}
& \multicolumn{2}{c|}{\textbf{ETTm1}}
& \multicolumn{2}{c|}{\textbf{ETTm2}}
& \multicolumn{2}{c|}{\textbf{ETTh1}}
& \multicolumn{2}{c}{\textbf{ETTh2}} \\
\cmidrule{2-9}
& MSE & MAE & MSE & MAE & MSE & MAE & MSE & MAE \\
\midrule
336
& 0.371 & 0.388
& 0.281 & 0.324
& \textbf{0.439} & \textbf{0.437}
& 0.387 & 0.404 \\

2000
& 0.357 & 0.384
& 0.264 & 0.324
& 0.447 & 0.445
& \textbf{0.353} & \textbf{0.391} \\

3000
& \textbf{0.347} & \textbf{0.379}
& \textbf{0.257} & \textbf{0.320}
& 0.515 & 0.481
& 0.360 & 0.401 \\

4000
& 0.362 & 0.386
& 0.263 & 0.322
& -- & --
& -- & -- \\

5000
& 0.380 & 0.395
& 0.266 & 0.321
& -- & --
& -- & -- \\
\bottomrule
\end{tabular}
}
\end{table}

\begin{table}[t]
\centering
\caption{Zero-shot forecasting performance of models trained on different synthetic data generators. 
All methods are evaluated on real-world ETT benchmarks without target-dataset fine-tuning. 
Lower MSE and MAE indicate better performance.}
\label{tab:zero_shot_synthetic_generators}
\resizebox{0.5\textwidth}{!}{
\begin{tabular}{l|cc|cc|cc|cc}
\toprule
\multirow{2}{*}{\textbf{Synthetic Generator}} 
& \multicolumn{2}{c|}{\textbf{ETTh1}} 
& \multicolumn{2}{c|}{\textbf{ETTh2}} 
& \multicolumn{2}{c|}{\textbf{ETTm1}} 
& \multicolumn{2}{c}{\textbf{ETTm2}} \\
\cmidrule{2-9}
& MSE $\downarrow$ & MAE $\downarrow$
& MSE $\downarrow$ & MAE $\downarrow$
& MSE $\downarrow$ & MAE $\downarrow$
& MSE $\downarrow$ & MAE $\downarrow$ \\
\midrule
CausalDynamics 
& 0.569 & 0.525
& \textbf{0.156} & 0.261
& 0.548 & 0.503
& 0.106 & 0.212 \\
LMC Synth 
& 0.789 & 0.571
& 0.179 & 0.279
& 0.690 & 0.506
& 0.093 & 0.204 \\
\textbf{Ours} 
& \textbf{0.425} & \textbf{0.427}
& 0.162 & \textbf{0.251}
& \textbf{0.512} & \textbf{0.474}
& \textbf{0.073} & \textbf{0.180} \\
\bottomrule
\end{tabular}
}
\end{table}

\section{Exploratory Zero-Shot Transfer with TS-SCM}
\label{sec:zero_shot_tsscm}

We further conduct an exploratory zero-shot transfer evaluation to examine whether TS-SCM-generated tasks can provide transferable forecasting priors. 
In this experiment, the forecasting model is trained only on synthetic tasks and directly evaluated on real-world ETT benchmarks without target-dataset fine-tuning. 
We emphasize that this setting is intended to assess the transfer potential of different synthetic data generation strategies, rather than to claim a complete solution to zero-shot time-series forecasting.

For evaluation, we construct each test instance using seven context samples, where each context sample contains a lookback window of length 96 and a future window of length 48. 
Given the target input sequence of length 96, the model is required to predict the next 48 time steps. 
We report the average MSE as the main evaluation metric. 
All compared models use the same evaluation protocol. 
We compare TS-SCM with two representative synthetic data generation strategies, including Causal Dynamics~\cite{herdeanu2025causaldynamics} and LMC Synth~\cite{taga2025timepfn}.

Table~\ref{tab:zero_shot_synthetic_generators} reports the zero-shot forecasting results on the ETT benchmarks. 
The proposed TS-SCM generator achieves the best MAE on all four datasets and the best MSE on three out of four datasets, including ETTh1, ETTm1, and ETTm2. 
On ETTh2, CausalDynamics obtains a slightly lower MSE, while TS-SCM still achieves the best MAE. 
These results suggest that the proposed forecasting-oriented synthetic task construction provides more transferable temporal priors than the compared synthetic generators in most evaluated settings.

Nevertheless, the advantage is not uniform across all metrics. 
In particular, the ETTh2 MSE result indicates that different synthetic priors may capture different aspects of real-world temporal dynamics. 
Therefore, we interpret these zero-shot results as encouraging evidence for the usefulness of TS-SCM-generated tasks, rather than as conclusive proof of a fully general zero-shot time-series forecasting model.

\section{Conclusion}

We presented Trio, a sample-aware time-series forecasting architecture that decomposes multivariate reasoning into temporal, spatial, and sample-level attention. By representing long histories as explicit lookback-future pairs, the model can retrieve historical input-output mappings for current prediction instead of treating history as a flat context. We further introduced TS-SCM, a forecasting-oriented synthetic task generator with dynamic lags, edge-level event scheduling, noise, feedback, and drift. Experiments on synthetic, industrial, and public benchmarks validate the effectiveness of the proposed architecture. Exploratory zero-shot results suggest that TS-SCM-generated tasks may provide useful transfer cues, but they do not yet establish a general-purpose PFN-style forecasting model. Future work will focus on stronger zero-shot evaluation, synthetic-data realism diagnostics, efficiency improvements, and broader comparisons with temporal causal priors.
\bibliography{aaai2026}

\appendix
\section{Appendix Overview}
This appendix provides additional materials that support the main paper. 
Appendix~\ref{sec:related work} presents extended discussions of related work, including deep learning methods for time-series forecasting, in-context learning, retrieval-augmented forecasting, and temporal causal priors. 
Appendix~\ref{sec:complexity} provides the complexity analysis of the proposed TSS encoder and compares the factorized attention design with flattened full attention.

We then provide more details about the proposed TS-SCM generator. 
Appendix~\ref{app:fractional_interpolation} explains the fractional interpolation strategy used for continuous time-shifted causal events, and Appendix~\ref{app:scm_examples} illustrates several hand-crafted SCM examples, including fixed-lag responses, variable-delay propagation, data drift, switching feedback, and multi-delay effects.

Finally, we report additional experimental results. 
Appendix~\ref{app:long_context_full} gives the full horizon-wise results of the long-context ablation study. 
We also include results on industrial datasets, an ablation study of the zero-gradient strategy, an ablation study of quantile-based context selection, and qualitative zero-shot visualizations comparing models trained with different synthetic data generation strategies. 
The appendix concludes with a discussion of current limitations and future directions.

\section{Related Works}
\label{sec:related work}

\subsection{Deep Learning in Time Series Forecasting}
In recent years, advances in machine learning have significantly accelerated progress in time series forecasting. Wu \emph{et al.}~\cite{wu2021autoformer} first proposed decomposing time series into trend and seasonal components and modeling periodic patterns efficiently via autocorrelation, enabling more stable long-horizon forecasting. Subsequently, inspired by Transformer architectures, Wu \emph{et al.}~\cite{Yuqietal2023PatchTST} explored partitioning time series into temporal blocks and applying self-attention to capture dependencies across blocks, achieving improved predictive performance. More recently, Wang et al. and Liu \emph{et al.}~\cite{liu2024itransformer} have incorporated exogenous variables to further enhance forecasting accuracy, while Wang \emph{et al.}~\cite{wang2024timexer} additionally investigated multi-scale representations to better capture temporal patterns at different resolutions.
Despite these advances in modeling temporal and spatial characteristics, most existing methods struggle to effectively exploit very long input sequences, which limits their performance in long-range forecasting scenarios.

\subsection{In-context Learning}
In recent years, several studies have begun to explore the use of in-context learning (ICL) to extend the amount of historical context available to time series models. Auer \emph{et al.}~\cite{auer25tirex} explored the use of xLSTM to enhance state-tracking capabilities in long sequences. Das \emph{et al.}~\cite{das2024a} proposed leveraging large-scale pretrained decoders to improve generalization in time series forecasting, while Lu \emph{et al.}~\cite{lu2025incontext} formulated the problem by constructing a series of (lookback, future) pairs within the input tokens.
However, these approaches primarily focus on increasing model capacity or expanding the scale of contextual inputs, without explicitly modeling the correspondence between lookback segments and their associated futures within historical data. As a result, historical information is often utilized as stacked context rather than as conditionally retrievable evidence for guiding current predictions.

\subsection{Retrieval-augmented and Example-based Time-Series Forecasting}

A related line of work enhances time-series forecasting by retrieving similar historical patterns or using previous segments as contextual evidence. Such methods are motivated by the observation that many real-world time series contain recurring motifs, delayed responses, and repeated local dynamics. Compared with purely parametric forecasting models, retrieval-based methods can explicitly reuse historical examples and are often effective when similar patterns reappear over time.

Our method shares the intuition that history should not be treated as a homogeneous sequence. However, instead of performing external nearest-neighbor search or retrieving only similar lookback windows, Trio organizes long histories into explicit lookback-future pairs. The future part of each historical pair is encoded as patch-wise prediction prompts, and the current future-query tokens retrieve information from these prompts through learned sample attention. This allows the model to learn soft, task-dependent retrieval in the representation space, rather than relying on a fixed distance metric over raw windows.

\subsection{Temporal Causal Priors and Synthetic Time-Series Generators}

Synthetic task generation is central to PFN-style learning. Existing time-series priors often rely on Gaussian-process kernels, linear coregionalization, VAR-like dynamics, or hand-crafted temporal patterns. These generators are useful for producing smooth or correlated sequences, but they do not always expose models to directed delayed mechanisms, state-dependent lags, feedback, and distributional changes.

Recent temporal causal simulators and benchmarks study richer dynamical systems, including structural causal priors, learned causal simulators, and causal discovery benchmarks. For example, causal-discovery-oriented benchmarks focus on recovering underlying temporal graphs from generated trajectories, while learned simulators aim to match real-data distributions through data-driven simulator selection. Our goal is different: TS-SCM is designed as a forecasting-oriented synthetic task prior. It directly constructs lookback-future episodes that match the input format of sample-level forecasting attention.

Compared with prior synthetic time-series generators, TS-SCM emphasizes three mechanisms. First, delayed effects are defined at the edge level instead of through a global lag matrix. Second, dynamic lags are implemented by a push-based event calendar, where parent effects are scheduled to future arrival times. Third, the generated trajectory is converted into structured forecasting episodes, allowing the forecasting model to learn from historical input-output mappings.

\section{Complexity Analysis}
\label{sec:complexity}

Let $B$ be the batch size, $V$ the number of variables, $\bar{S}=S+1$ the number of windows including the current window, $P=P_x+P_y$ the number of tokens per window, and $D$ the hidden dimension. A full attention operation over all variable, sample, and patch dimensions would require attention over $V\bar{S}P$ tokens, resulting in
\begin{equation}
\mathcal{O}\!\left(B(V\bar{S}P)^2D\right),
\end{equation}
which is prohibitive for long-context forecasting.

The proposed TSS encoder factorizes this operation into three axis-wise attention modules. The spatial module attends over variables:
\begin{equation}
\mathcal{O}\!\left(B\bar{S}P V^2D\right).
\end{equation}
The temporal module attends over patch tokens:
\begin{equation}
\mathcal{O}\!\left(BV\bar{S}P^2D\right).
\end{equation}
The sample module attends over historical windows for each variable and future-query position:
\begin{equation}
\mathcal{O}\!\left(BVP_yS^2D\right).
\end{equation}
Thus, one TSS block has complexity
\begin{equation}
\mathcal{O}\!\left(
B\bar{S}PV^2D
+
BV\bar{S}P^2D
+
BVP_yS^2D
\right).
\end{equation}

This factorization avoids the quadratic cost of flattened full attention over all axes. Nevertheless, the sample module still scales quadratically with the number of historical windows $S$. In the current implementation, we keep $S$ moderate and use the zero-gradient strategy to reduce memory usage. For much longer histories, top-$k$ sample selection, memory compression, or low-rank sample attention could be incorporated to further reduce the $S^2$ cost. We leave these scalable retrieval variants for future work.

\section{More information about TS-SCM}
\subsection{Interpolation for Fractional Time-shifted Events}
\label{app:fractional_interpolation}

In our synthetic causal time-series generation process, a child variable may receive a delayed or shifted signal from its parent variable. 
When the time shift is continuous, the shifted arrival position is not necessarily aligned with an integer time index. 
A naive implementation would round the arrival position to the nearest integer time step. 
However, such a rounding operation may introduce artificial discontinuities and temporal misalignment, especially when the delay changes over time. 
To avoid this issue, we use a fractional interpolation strategy to redistribute the parent signal onto integer time bins.

Specifically, let the fractional arrival position at time step $t$ be
\begin{equation}
    s(t) = t + d_t,
\end{equation}
where $d_t$ denotes the time-varying delay. 
For two consecutive parent observations, we record their fractional arrival positions and values as
\begin{equation}
    s_0 = (t-1) + d_{t-1}, 
    \qquad 
    x_0 = x_{\mathrm{parent}}[t-1],
\end{equation}
and
\begin{equation}
    s_1 = t + d_t, 
    \qquad 
    x_1 = x_{\mathrm{parent}}[t].
\end{equation}

If the interval $[s_0, s_1]$ crosses one or more integer time bins, we assign events to these crossed integer positions. 
For each crossed integer position $k$, we compute its relative location between $s_0$ and $s_1$:
\begin{equation}
    \beta = \frac{k - s_0}{s_1 - s_0}, 
    \qquad \beta \in [0,1].
\end{equation}
The corresponding parent value at position $k$ is then obtained by linear interpolation:
\begin{equation}
    \hat{x}_k = (1-\beta)x_0 + \beta x_1.
\end{equation}
This interpolated value is used as the event value arriving at the integer time bin $k$.

When $s_1 > s_0$, the crossed integer positions are given by
\begin{equation}
    k \in \left\{ \lfloor s_0 \rfloor + 1, \ldots, \lfloor s_1 \rfloor \right\}.
\end{equation}
When $s_1 < s_0$, which can occur under abrupt changes in the time-varying delay, the same procedure is applied in the reverse direction. 
This ensures that all integer bins crossed by the fractional trajectory are considered.

Compared with direct rounding, the proposed interpolation strategy better preserves the continuous temporal structure of the shifted parent signal. 
It avoids assigning a delayed event to an arbitrary nearest bin and instead distributes events according to the actual fractional trajectory. 
As a result, the generated child sequence can better reflect smooth delayed responses, phase shifts, and non-stationary temporal propagation patterns.

\subsection{Illustrative Examples of TS-SCM Dynamics.}~\label{app:scm_examples}

\begin{figure*}[t]
    \centering
    \includegraphics[width=\textwidth]{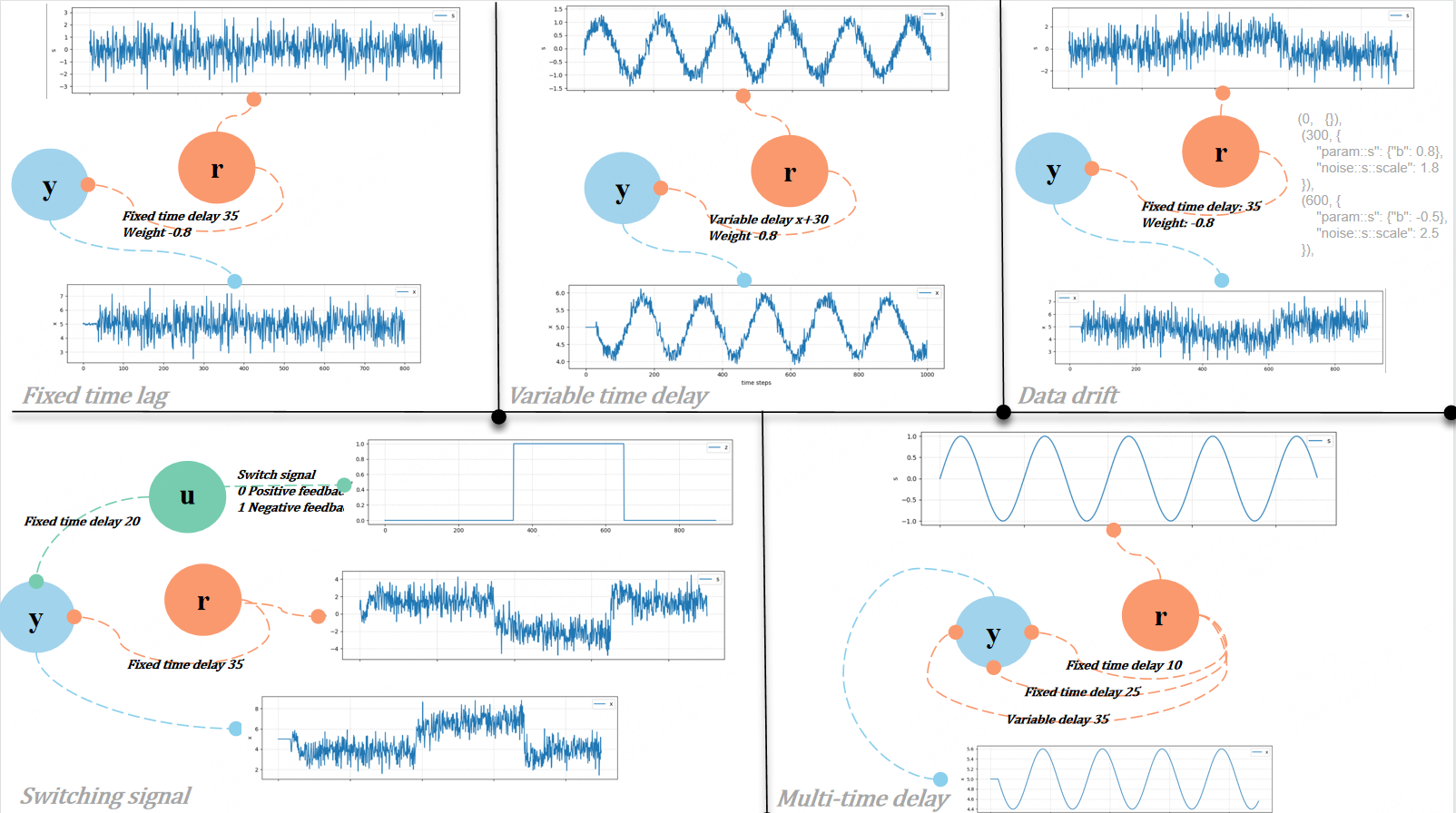}
    \caption{
    Illustrative examples of TS-SCM-generated structural causal dynamics. 
    Each panel shows a hand-crafted structural causal graph together with the generated time series. 
    By configuring edge-level delay, weight, feedback, switching signals, and distribution drift, TS-SCM can produce diverse temporal patterns such as fixed-lag responses, variable-delay propagation, data drift, switching feedback, and multi-delay effects.
    }
    \label{fig:scm_examples}
\end{figure*}

To illustrate how TS-SCM constructs diverse synthetic forecasting tasks, we provide several hand-crafted examples of structural causal dynamics in Fig.~\ref{fig:scm_examples}. 
Each example consists of a small structural causal graph and the corresponding generated time series. 
The nodes represent different variables, while directed edges specify temporal causal dependencies with configurable delay, weight, and transformation functions. 
By changing these edge-level mechanisms, TS-SCM can generate diverse temporal patterns, including fixed-lag responses, variable-delay propagation, distribution drift, switching feedback, and multi-delay causal effects.

In the fixed-lag example, the target variable receives a delayed signal from its parent variable with a constant time lag and edge weight. 
This setting simulates common delayed-response patterns where the effect of one variable appears after a fixed number of time steps. 
The variable-delay example further allows the delay to change over time, producing non-stationary phase shifts between the parent and child variables. 
The data-drift example introduces changes in the underlying data-generating parameters, such as noise scale or distributional parameters, leading to non-stationary temporal behavior. 
The switching-signal example uses an additional control variable to switch between different feedback regimes, allowing the generated sequence to exhibit regime-dependent dynamics. 
Finally, the multi-delay example combines multiple delayed causal paths between variables, producing richer temporal dependencies with both short-term and long-term effects.

These examples show that TS-SCM does not rely on a single fixed synthetic pattern. 
Instead, it defines synthetic tasks through compositional structural mechanisms, where different temporal dependencies can be created by combining causal edges, time delays, feedback, switching signals, and distribution shifts. 
Such diversity is important for exposing the forecasting model to a broader range of temporal structures during synthetic pretraining.

\section{More Experimental Results}
\subsection{Results on Industrial Datasets}

\begin{table}[t]
\centering
\caption{Performance comparison (MSE and MAE) on A, B, and C datasets. Bold indicates the best result.}
\label{tab:methods_comparison_3dec}

\scalebox{0.8}{

\begin{tabular}{c|cc|cc|cc}
\toprule
\multirow{2}{*}{Methods} & \multicolumn{2}{c|}{\textbf{A}} & \multicolumn{2}{c|}{\textbf{B}} & \multicolumn{2}{c}{\textbf{C}} \\
\cmidrule{2-7}
 & MSE & MAE & MSE & MAE & MSE & MAE \\
\midrule
TimesNet      & 0.090 & 0.130 & 0.129 & 0.120 & 0.086 & 0.167 \\
PatchTST      & 0.090 & 0.144 & 0.122 & 0.111 & 0.120 & 0.174 \\
iTransformer  & 0.092 & 0.132 & 0.140 & 0.115 & 0.072 & 0.151 \\
TimeXer       & 0.131 & 0.154 & 0.134 & 0.114 & 0.085 & 0.168 \\
PAttn         & 0.088 & 0.136 & 0.124 & 0.109 & 0.118 & 0.174 \\
\midrule
Baseline      & 0.076 & 0.117 & 0.134 & \textbf{0.109} & 0.068 & 0.143 \\
\rowcolor{gray!30} 
Trio & \textbf{0.072} & \textbf{0.116} & \textbf{0.117} & 0.111 & \textbf{0.066} & \textbf{0.140} \\
\bottomrule
\end{tabular}
}
\end{table}

Table~\ref{tab:methods_comparison_3dec} reports the results on three industrial datasets.
Compared with existing forecasting models, the proposed Trio model achieves the best MSE on all three datasets.
It also obtains the best MAE on A and B.
Compared with the base model, Trio reduces the MSE from 0.076 to 0.072 on A, from 0.134 to 0.117 on B, and from 0.068 to 0.066 on C.

These results indicate that explicitly exploiting historical input--output correspondences is beneficial for real industrial time series.
This is consistent with the characteristics of industrial systems, where the current state often depends on delayed responses, long-range temporal patterns, and interactions among multiple variables.
Although the MAE improvement is not always large, the consistent MSE reduction suggests that the proposed architecture is effective at reducing large prediction deviations.

\begin{table}[t]
\centering
\caption{Comparison with GTR and RAFT on overlapping benchmark datasets. Metric is MSE; lower is better.}
\label{tab:trio_gtr_raft_comparison}

\small
\setlength{\tabcolsep}{6pt}
\renewcommand{\arraystretch}{1.08}

\begin{tabular}{lccc}
\toprule
Dataset & \textbf{Trio} & GTR & RAFT \\
\midrule
ETTm1
& \textbf{0.358} & 0.367 & 0.381 \\

ETTm2
& \textbf{0.255} & 0.268 & 0.281 \\

Electricity
& \textbf{0.164} & 0.166 & 0.175 \\

Weather
& 0.249 & \textbf{0.239} & 0.270 \\
\midrule
Count
& \textbf{3} & 1 & 0 \\
\bottomrule
\end{tabular}
\end{table}

\subsection{Compared with the retrieval-based methods}
Table~\ref{tab:trio_gtr_raft_comparison} compares Trio with GTR~\cite{cao2026enhancing} and RAFT~\cite{tire2024retrieval} on the overlapping benchmark datasets. 
Trio achieves the best MSE on three out of four datasets, including ETTm1, ETTm2, and Electricity, while GTR performs better on Weather. 
These results suggest that Trio is competitive with recent retrieval-augmented forecasting methods, especially on ETT and electricity forecasting benchmarks.

\subsection{Full Results of Long-Context Ablation}
\label{app:long_context_full}

The averaged results are summarized in Tables~\ref{tab:long_context_avg} and~\ref{tab:long_context_avg_336} in the main text, while the complete horizon-wise results are reported here. 
Tables~\ref{tab:long_context_full} and~\ref{tab:long_context_full_336} correspond to the default input lengths of 96 and 336, respectively. 
For each setting, we vary the context length $L$ and evaluate the model under prediction horizons $H \in \{96, 192, 336, 720\}$.

As shown in Table~\ref{tab:long_context_full}, when the default input length is 96, increasing the context length generally improves the performance compared with the default setting. 
However, the best context length varies across datasets and prediction horizons. 
For instance, moderate context lengths such as $L=1000$ or $L=2000$ often achieve strong performance on ETTm1, ECL, ETTh1, and ETTh2, while longer contexts are more beneficial in some ETTm2 and Weather settings. 
This suggests that historical information beyond the default input window can provide useful temporal cues, but the optimal context length is dataset-dependent.

Table~\ref{tab:long_context_full_336} further reports the results when the default input length is 336. 
Compared with the default setting, longer contexts also improve the average performance on ETTm1 and ETTm2, with $L=3000$ achieving the best average results on these two datasets. 
Nevertheless, further increasing the context length to $L=4000$ or $L=5000$ does not consistently bring additional gains and may even degrade the performance. 
One possible reason is that longer input windows reduce the number of available training samples under a fixed-length time series, which may weaken the effective supervision during training. 
Moreover, excessively long contexts may introduce redundant or noisy historical patterns, making it harder for the model to focus on the most relevant temporal dependencies.

Overall, these full results support the conclusion in the main text: long-context modeling is beneficial, but simply increasing the context length is not always optimal. 
A moderate context length can better balance useful historical information, training sample availability, and noise introduced by overly long inputs.

\begin{table}[t]
\centering
\caption{Full performance comparison across different context lengths ($L$) and prediction horizons ($H$) with the default input length of 96. Lower values indicate better performance}
\label{tab:long_context_full}

\scriptsize
\setlength{\tabcolsep}{2.6pt}
\renewcommand{\arraystretch}{1.03}

\resizebox{0.5\textwidth}{!}{
\begin{tabular}{c|cc|cc|cc|cc|cc|cc}
\toprule
\multirow{2}{*}{Context Length ($L$)}
& \multicolumn{2}{c|}{\textbf{ETTm1}}
& \multicolumn{2}{c|}{\textbf{ETTm2}}
& \multicolumn{2}{c|}{\textbf{Weather}}
& \multicolumn{2}{c|}{\textbf{ECL}}
& \multicolumn{2}{c|}{\textbf{ETTh1}}
& \multicolumn{2}{c}{\textbf{ETTh2}} \\
\cmidrule{2-13}
& MSE & MAE & MSE & MAE & MSE & MAE & MSE & MAE & MSE & MAE & MSE & MAE \\
\midrule

\multicolumn{13}{c}{\cellcolor{gray!15}\textbf{Prediction Horizon $H = 96$}} \\
\midrule
96
& 0.313 & 0.350
& 0.173 & 0.251
& 0.157 & 0.198
& 0.142 & 0.235
& 0.386 & 0.396
& 0.287 & \textbf{0.334} \\
1000
& \textbf{0.289} & \textbf{0.336}
& 0.159 & \textbf{0.246}
& \textbf{0.149} & \textbf{0.193}
& \textbf{0.133} & \textbf{0.227}
& \textbf{0.374} & \textbf{0.393}
& 0.281 & \textbf{0.334} \\
2000
& \textbf{0.289} & 0.341
& \textbf{0.156} & 0.249
& 0.154 & 0.203
& 0.134 & 0.231
& 0.388 & 0.398
& \textbf{0.279} & 0.336 \\
3000
& 0.312 & 0.349
& 0.155 & 0.244
& 0.156 & 0.204
& 0.139 & 0.235
& 0.484 & 0.403
& -- & -- \\
4000
& 0.294 & 0.350
& \textbf{0.156} & 0.247
& 0.161 & 0.209
& 0.141 & 0.238
& -- & --
& -- & -- \\

\midrule
\multicolumn{13}{c}{\cellcolor{gray!15}\textbf{Prediction Horizon $H = 192$}} \\
\midrule
96
& 0.366 & 0.379
& 0.240 & 0.297
& 0.217 & 0.255
& 0.159 & 0.250
& 0.440 & 0.427
& 0.367 & 0.384 \\
1000
& 0.344 & \textbf{0.368}
& 0.232 & 0.298
& 0.212 & 0.254
& \textbf{0.156} & \textbf{0.248}
& \textbf{0.422} & \textbf{0.422}
& 0.361 & \textbf{0.388} \\
2000
& 0.339 & 0.371
& 0.219 & 0.293
& 0.210 & \textbf{0.246}
& 0.162 & 0.256
& 0.433 & 0.427
& \textbf{0.358} & 0.391 \\
3000
& 0.337 & 0.372
& 0.214 & \textbf{0.288}
& 0.203 & 0.250
& 0.162 & 0.256
& -- & --
& -- & -- \\
4000
& \textbf{0.333} & 0.374
& \textbf{0.212} & \textbf{0.288}
& \textbf{0.202} & 0.253
& 0.163 & 0.256
& --4 & --
& -- & -- \\

\midrule
\multicolumn{13}{c}{\cellcolor{gray!15}\textbf{Prediction Horizon $H = 336$}} \\
\midrule
96
& 0.399 & 0.399
& 0.300 & 0.334
& 0.275 & 0.296
& 0.173 & \textbf{0.265}
& 0.478 & 0.446
& 0.418 & 0.425 \\
1000
& 0.371 & \textbf{0.389}
& 0.291 & 0.337
& 0.270 & \textbf{0.293}
& \textbf{0.172} & 0.266
& \textbf{0.466} & \textbf{0.440}
& \textbf{0.380} & \textbf{0.407} \\
2000
& 0.377 & 0.402
& 0.285 & 0.340
& 0.281 & 0.305
& 0.177 & 0.276
& 0.476 & 0.447
& 0.400 & 0.429 \\
3000
& 0.368 & 0.393
& 0.273 & 0.330
& 0.272 & 0.301
& 0.173 & 0.271
& 0.484 & 0.457
& -- & -- \\
4000
& \textbf{0.365} & 0.392
& \textbf{0.272} & \textbf{0.329}
& \textbf{0.261} & 0.294
& 0.180 & 0.276
& 0.498 & 0.464
& -- & -- \\

\midrule
\multicolumn{13}{c}{\cellcolor{gray!15}\textbf{Prediction Horizon $H = 720$}} \\
\midrule
96
& 0.469 & 0.438
& 0.409 & 0.399
& 0.358 & 0.348
& \textbf{0.194} & \textbf{0.286}
& \textbf{0.494} & \textbf{0.465}
& 0.432 & 0.443 \\
1000
& 0.419 & \textbf{0.419}
& 0.370 & 0.392
& 0.364 & 0.352
& 0.195 & 0.288
& 0.497 & 0.483
& \textbf{0.392} & \textbf{0.426} \\
2000
& 0.428 & 0.428
& \textbf{0.360} & \textbf{0.389}
& 0.347 & \textbf{0.345}
& 0.197 & 0.292
& 0.610 & 0.538
& 0.427 & 0.461 \\
3000
& \textbf{0.418} & 0.425
& 0.384 & 0.399
& 0.356 & 0.352
& 0.199 & 0.294
& -- & --
& -- & -- \\
4000
& 0.469 & 0.457
& 0.425 & 0.416
& \textbf{0.346} & 0.355
& 0.205 & 0.301
& -- & --
& -- & -- \\

\bottomrule
\end{tabular}
}
\end{table}

\begin{table}[t]
\centering
\caption{Full performance comparison across different context lengths ($L$) and prediction horizons ($H$) with the default input length of 336. Lower values indicate better performance.}
\label{tab:long_context_full_336}

\scriptsize
\setlength{\tabcolsep}{4pt}
\renewcommand{\arraystretch}{1.03}

\resizebox{0.5\textwidth}{!}{
\begin{tabular}{c|cc|cc|cc|cc}
\toprule
\multirow{2}{*}{Context Length ($L$)}
& \multicolumn{2}{c|}{\textbf{ETTm1}}
& \multicolumn{2}{c|}{\textbf{ETTm2}}
& \multicolumn{2}{c|}{\textbf{ETTh1}}
& \multicolumn{2}{c}{\textbf{ETTh2}} \\
\cmidrule{2-9}
& MSE & MAE & MSE & MAE & MSE & MAE & MSE & MAE \\
\midrule

\multicolumn{9}{c}{\cellcolor{gray!15}\textbf{Prediction Horizon $H = 96$}} \\
\midrule
336
& 0.299 & 0.349
& 0.164 & 0.250
& 0.386 & \textbf{0.399}
& \textbf{0.280} & 0.337 \\

2000
& 0.298 & 0.344
& 0.162 & 0.254
& \textbf{0.380} & 0.400
& 0.281 & \textbf{0.336} \\

3000
& \textbf{0.278} & \textbf{0.335}
& 0.161 & 0.253
& 0.381 & 0.400
& 0.289 & 0.349 \\

4000
& 0.286 & 0.338
& \textbf{0.157} & \textbf{0.247}
& -- & --
& -- & -- \\

5000
& 0.289 & 0.338
& \textbf{0.157} & 0.251
& -- & --
& -- & -- \\

\midrule
\multicolumn{9}{c}{\cellcolor{gray!15}\textbf{Prediction Horizon $H = 192$}} \\
\midrule
336
& 0.351 & 0.385
& 0.276 & 0.315
& \textbf{0.426} & \textbf{0.429}
& 0.400 & 0.406 \\

2000
& \textbf{0.323} & \textbf{0.364}
& 0.227 & 0.298
& 0.439 & 0.439
& 0.347 & \textbf{0.382} \\

3000
& 0.325 & 0.365
& \textbf{0.219} & 0.294
& 0.428 & \textbf{0.429}
& \textbf{0.345} & 0.385 \\

4000
& 0.329 & \textbf{0.364}
& 0.221 & \textbf{0.293}
& -- & --
& -- & -- \\

5000
& 0.336 & 0.367
& 0.222 & \textbf{0.293}
& -- & --
& -- & -- \\

\midrule
\multicolumn{9}{c}{\cellcolor{gray!15}\textbf{Prediction Horizon $H = 336$}} \\
\midrule
336
& 0.378 & 0.390
& 0.307 & 0.341
& \textbf{0.464} & 0.452
& 0.448 & 0.433 \\

2000
& 0.374 & 0.396
& 0.283 & 0.338
& 0.465 & \textbf{0.451}
& \textbf{0.377} & \textbf{0.404} \\

3000
& 0.367 & \textbf{0.389}
& 0.277 & 0.337
& 0.471 & 0.459
& 0.402 & 0.426 \\

4000
& \textbf{0.366} & 0.395
& \textbf{0.276} & \textbf{0.334}
& -- & --
& -- & -- \\

5000
& 0.388 & 0.399
& 0.286 & 0.336
& -- & --
& -- & -- \\

\midrule
\multicolumn{9}{c}{\cellcolor{gray!15}\textbf{Prediction Horizon $H = 720$}} \\
\midrule
336
& 0.455 & 0.428
& 0.376 & \textbf{0.390}
& \textbf{0.479} & \textbf{0.468}
& 0.418 & \textbf{0.438} \\

2000
& 0.432 & 0.432
& 0.384 & 0.405
& 0.502 & 0.491
& 0.407 & 0.441 \\

3000
& \textbf{0.418} & \textbf{0.426}
& \textbf{0.372} & 0.397
& 0.778 & 0.635
& \textbf{0.406} & 0.445 \\

4000
& 0.465 & 0.445
& 0.398 & 0.414
& -- & --
& -- & -- \\

5000
& 0.509 & 0.474
& 0.399 & 0.405
& -- & --
& -- & -- \\

\bottomrule
\end{tabular}
}
\end{table}

\subsection{Discussion on the Zero-Gradient Strategy}
The zero-gradient strategy is used only in dataset-specific supervised training with long historical contexts. 
In this setting, the prediction loss is defined on the current future sequence, while historical windows mainly serve as contextual evidence. 
Therefore, we process historical windows without gradient tracking and recompute the current window with gradients, which reduces memory consumption and stabilizes optimization.

It is important to note that this strategy is not a core component of the proposed sample-attention mechanism, nor is it required by the TS-SCM generator. 
Instead, it is a practical training option for long-context supervised forecasting on specific datasets. 
When the model is trained or evaluated under other settings, such as synthetic-task training or exploratory zero-shot transfer, the use of zero-gradient processing can be disabled or adjusted according to the training objective.

This strategy may limit the adaptation of historical representations during each update, and should therefore be viewed as an efficiency-oriented approximation rather than a theoretically optimal design. 
Empirically, Table~\ref{tab:ablation_context_len} shows that it generally improves or maintains performance under long-context supervised training. 
Exploring partial-gradient alternatives, such as lightweight adapters, memory compression, or curriculum-based gradient schedules for historical windows, is left for future work.

Table~\ref{tab:ablation_context_len} studies the zero-gradient strategy used for historical windows.
This strategy treats historical windows mainly as conditional context and retains gradients only for the current prediction window.
The results show that \texttt{zero\_grad} generally improves or maintains performance, especially when the context length becomes large.
For example, on ETTm1 with context length 3000, the MSE decreases from 0.362 to 0.360.
On ETTm2 with context length 3000, the MSE decreases from 0.265 to 0.262.
These results suggest that the zero-gradient strategy can reduce optimization instability while still allowing the model to benefit from long historical information.

\begin{table}[htbp]
\centering
\caption{Ablation study on the optional \texttt{zero\_grad} strategy for dataset-specific long-context supervised training on ETTm1 and ETTm2.}
\label{tab:ablation_context_len}
\scalebox{0.6}{   
\begin{tabular}{c|cc|cc|cc|cc}
\toprule
\multirow{3}{*}{Context Length} & \multicolumn{4}{c|}{ETTm1} & \multicolumn{4}{c}{ETTm2} \\
\cline{2-9}
 & \multicolumn{2}{c|}{w/o \texttt{zero\_grad}} & \multicolumn{2}{c|}{w \texttt{zero\_grad}} & \multicolumn{2}{c|}{w/o \texttt{zero\_grad}} & \multicolumn{2}{c}{w \texttt{zero\_grad}} \\
\cline{2-9}
 & MSE$\downarrow$  & MAE$\downarrow$  & MSE$\downarrow$  & MAE$\downarrow$  & MSE$\downarrow$  & MAE$\downarrow$  & MSE$\downarrow$  & MAE$\downarrow$  \\
\midrule
1000 & \textbf{0.371} & 0.405 & 0.372 & 0.405 & 0.272 & 0.333 & \textbf{0.271} & \textbf{0.331} \\
2000 & 0.372 & 0.407 & \textbf{0.367} & \textbf{0.403} & 0.277 & 0.339 & \textbf{0.265} & \textbf{0.331} \\
3000 & 0.362 & 0.405 & \textbf{0.360} & \textbf{0.405} & 0.265 & 0.332 & \textbf{0.262} & \textbf{0.329} \\
4000 & 0.377 & 0.415 & \textbf{0.368} & \textbf{0.409} & \textbf{0.272} & 0.342 & 0.279 & \textbf{0.339} \\
\bottomrule
\end{tabular}%
}
\end{table}

\subsection{Ablation on Quantile-based Context Selection.}

\begin{table}[t]
\centering
\caption{Ablation study of quantile-based context selection and cross-attention under the prediction length of 96. Lower values indicate better performance.}
\label{tab:quantile_ablation}

\scriptsize
\setlength{\tabcolsep}{4.5pt}
\renewcommand{\arraystretch}{1.08}

\resizebox{0.5\textwidth}{!}{
\begin{tabular}{c c|cc|cc|cc|cc}
\toprule
\multirow{2}{*}{Selection}
& \multirow{2}{*}{Method}
& \multicolumn{2}{c|}{\textbf{ETTm1}}
& \multicolumn{2}{c|}{\textbf{ETTm2}}
& \multicolumn{2}{c|}{\textbf{Weather}}
& \multicolumn{2}{c}{\textbf{ECL}} \\
\cmidrule{3-10}
& 
& MSE & MAE
& MSE & MAE
& MSE & MAE
& MSE & MAE \\
\midrule

\multirow{2}{*}{Quantile}
& Baseline
& 0.387 & 0.391
& 0.281 & 0.320
& 0.252 & \textbf{0.274}
& 0.158 & 0.250 \\

& Cross-Attention
& \textbf{0.358} & \textbf{0.385}
& \textbf{0.255} & \textbf{0.318}
& \textbf{0.248} & 0.275
& \textbf{0.154} & \textbf{0.247} \\
\midrule

\multirow{2}{*}{MSE}
& Baseline
& 0.376 & \textbf{0.396}
& 0.280 & 0.335
& 0.272 & 0.293
& \textbf{0.176} & \textbf{0.272} \\

& Cross-Attention
& \textbf{0.367} & 0.403
& \textbf{0.264} & \textbf{0.327}
& \textbf{0.249} & \textbf{0.288}
& 0.180 & 0.279 \\

\bottomrule
\end{tabular}
}
\end{table}

To analyze the effect of the proposed quantile-based context selection strategy, we compare it with an MSE-based selection strategy under the prediction length of 96. 
For each selection strategy, the baseline directly uses the selected historical contexts, while the Cross-Attention variant further integrates the selected contexts with the current input sequence through cross-attention.

As shown in Table~\ref{tab:quantile_ablation}, the quantile-based strategy achieves consistent improvements after introducing cross-attention on most datasets. 
For example, the MSE is reduced from 0.387 to 0.358 on ETTm1, from 0.281 to 0.255 on ETTm2, from 0.158 to 0.154 on ECL, from 0.450 to 0.440 on ETTh1, and from 0.376 to 0.353 on ETTh2. 
This demonstrates that quantile-based context selection can provide informative historical patterns, and cross-attention further helps the model exploit these selected contexts.

Compared with the MSE-based selection strategy, the quantile-based strategy shows more stable performance. 
Although the MSE-based strategy can also benefit from cross-attention on ETTm1, ETTm2, and Weather, its improvement is less consistent, and the performance on ECL slightly degrades after adding cross-attention. 
This suggests that directly selecting contexts according to point-wise MSE may overemphasize local numerical similarity, whereas quantile-based selection can better capture distribution-level temporal characteristics.

\subsection{Visualization}~\label{vis}
\begin{figure*}[t]
    \centering
    \includegraphics[width=0.8\linewidth]{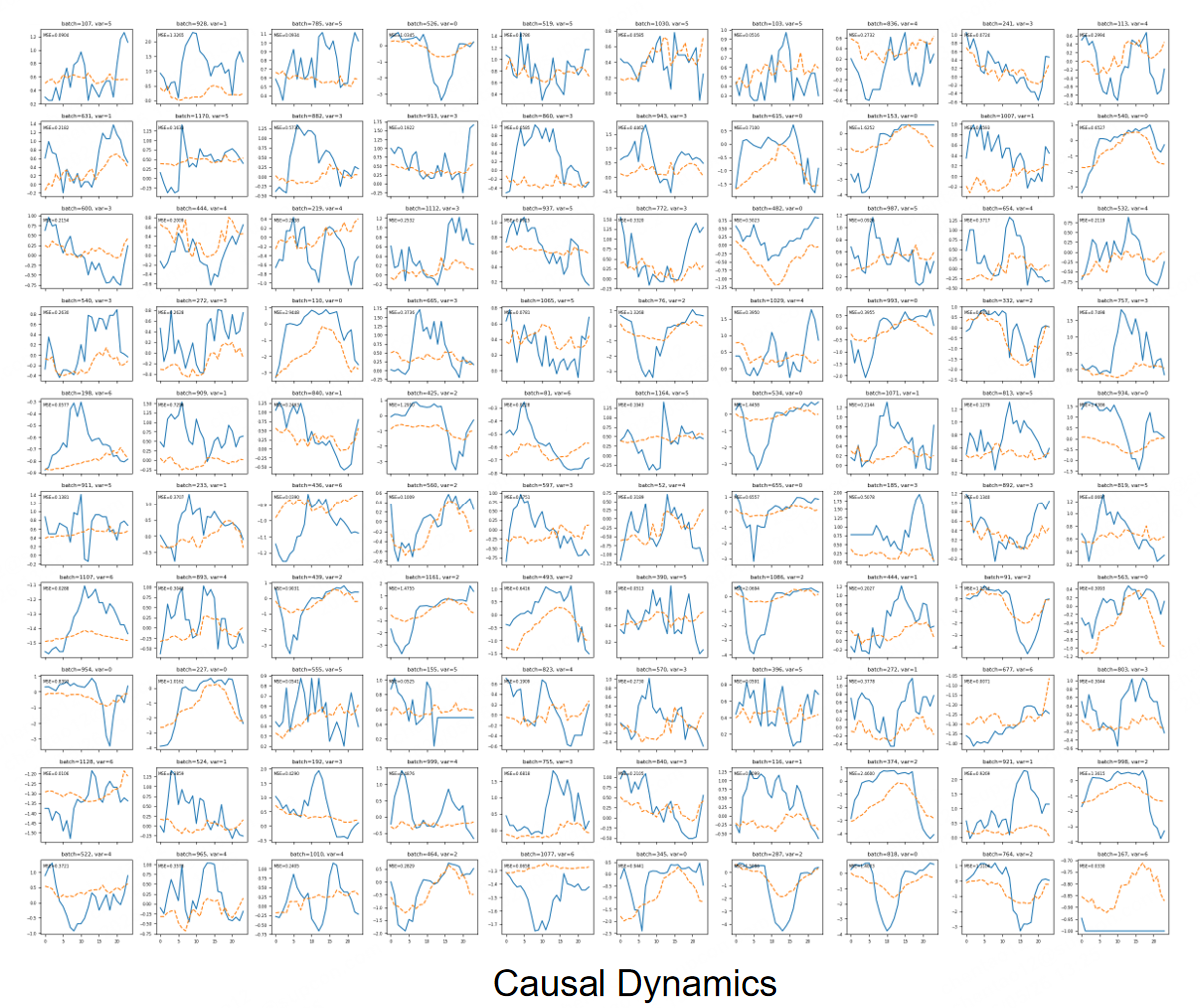}
    \caption{
    Qualitative results of the model trained with Causal Dynamics synthetic data.
    The blue curves denote the ground-truth sequences, and the orange dashed curves denote model predictions.
    }
    \label{fig:causal_dyn}
\end{figure*}

\begin{figure*}[t]
    \centering
    \includegraphics[width=0.8\linewidth]{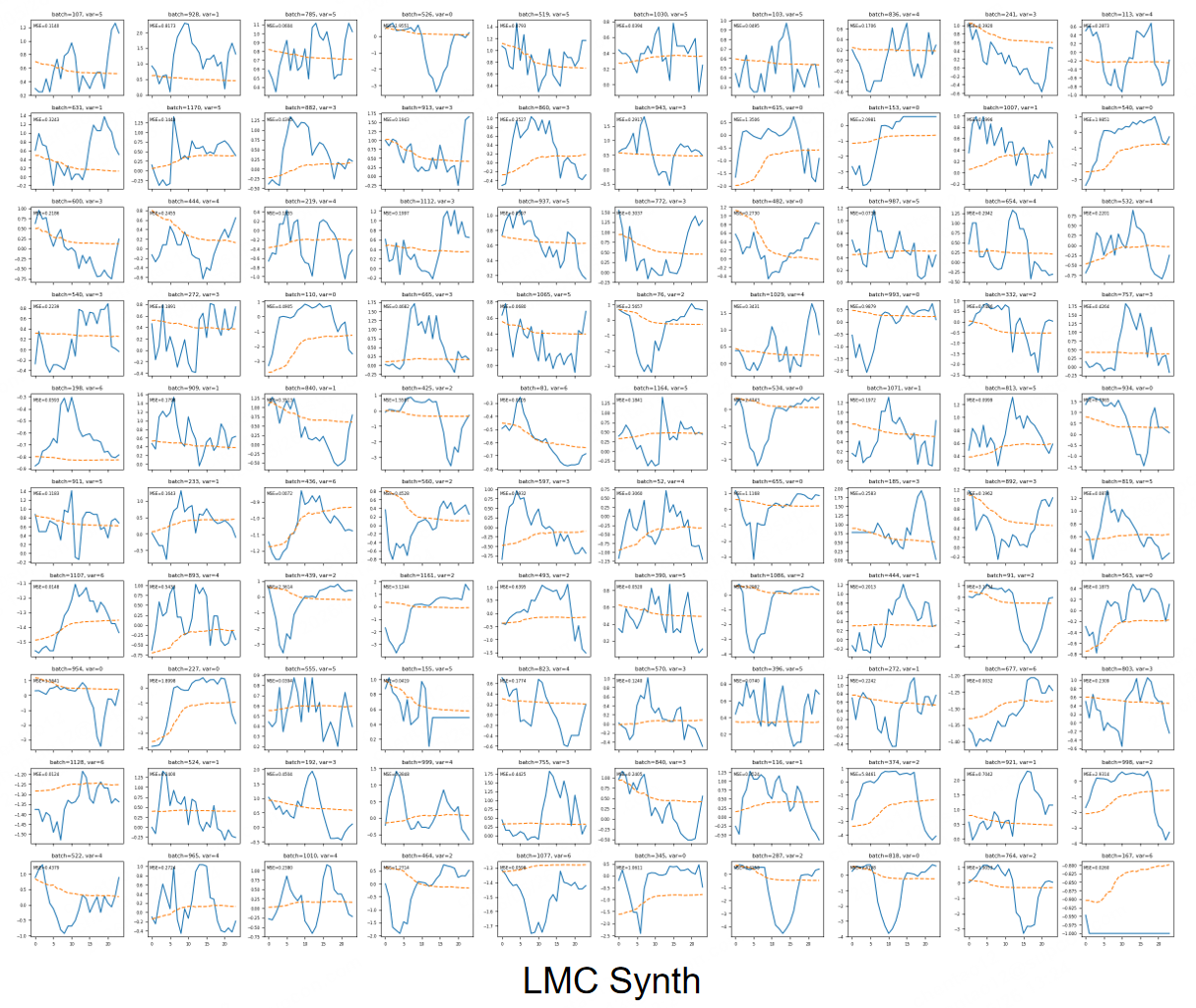}
    \caption{
    Qualitative results of the model trained with LMC Synth synthetic data.
    The blue curves denote the ground-truth sequences, and the orange dashed curves denote model predictions.
    }
    \label{fig:lmc_synth}
\end{figure*}

\begin{figure*}[t]
    \centering
    \includegraphics[width=0.8\linewidth]{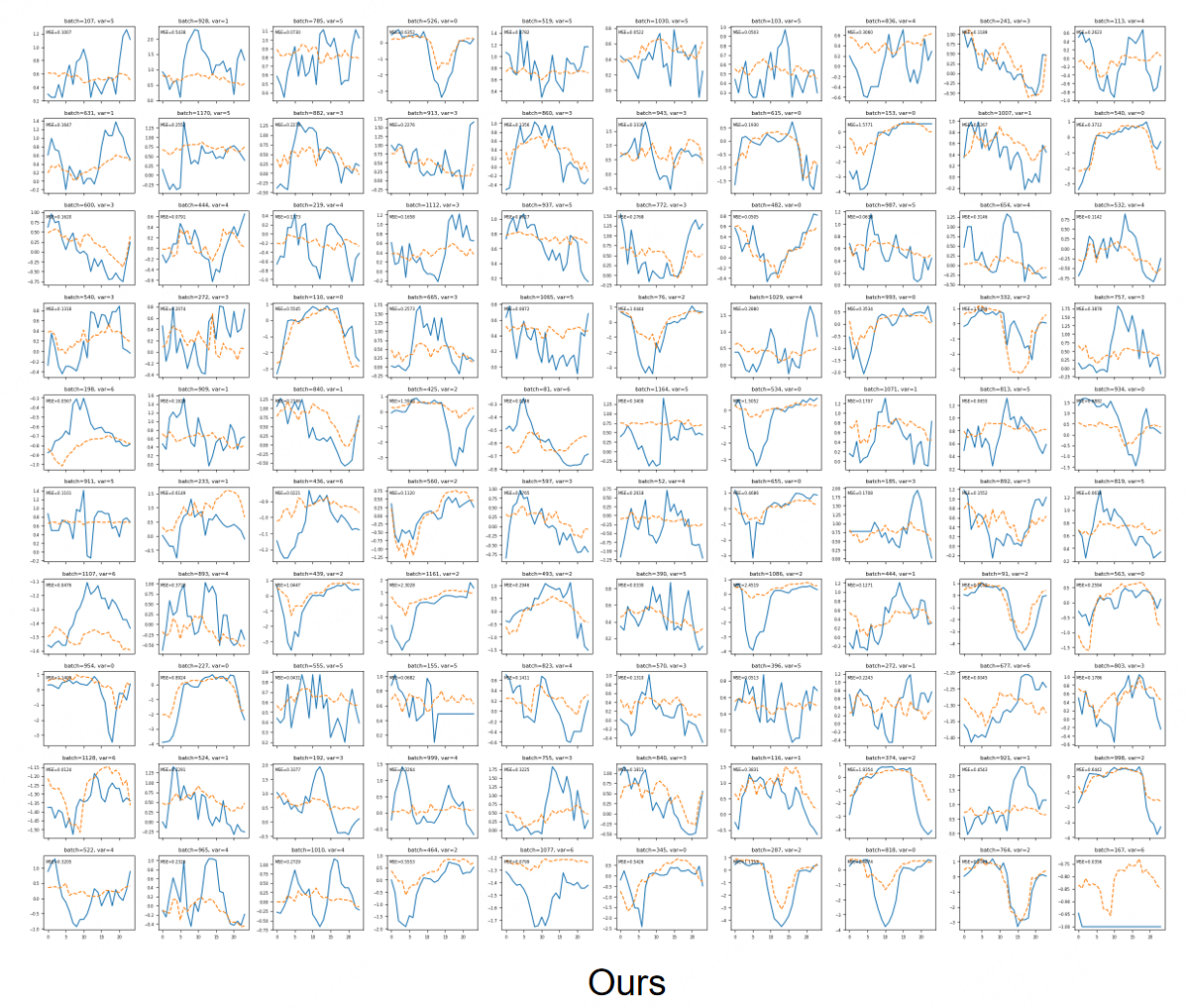}
    \caption{
    Qualitative results of the model trained with our synthetic data.
    The blue curves denote the ground-truth sequences, and the orange dashed curves denote model predictions.
    }
    \label{fig:ours_synth}

\end{figure*}

Figures~\ref{fig:causal_dyn}--\ref{fig:ours_synth} provide a qualitative comparison of models trained with different synthetic data generation strategies. 
Causal Dynamics mainly captures coarse global trends but often fails to reproduce local fluctuations and sharp turning points. 
LMC Synth produces more stable predictions, yet its outputs are generally over-smoothed and tend to miss high-frequency variations. 
In contrast, the model trained with our synthetic data better follows local temporal variations and preserves peak-valley structures, indicating stronger shape-level generalization. 
Nevertheless, sharp extrema remain challenging, as the predictions are still relatively conservative around abrupt changes.

\section{Limitations}

This work has several limitations. First, although the proposed framework is motivated by PFN-style learning, most main experiments are still conducted under supervised forecasting settings. The TS-SCM-based zero-shot results are preliminary and should be interpreted as evidence of transfer potential rather than a complete solution to general-purpose time-series forecasting.

Second, the zero-gradient strategy for historical windows is an empirical optimization design. While it reduces memory cost and stabilizes training, it may also limit the adaptation of historical representations. More flexible strategies, such as partial gradients, adapters, or curriculum-based historical updates, deserve further investigation.

Third, the current TS-SCM generator is hand-designed. Although it supports dynamic lags, event calendars, drift, and heterogeneous node mechanisms, its realism is not fully validated against real-world time-series distributions. Future work should include stronger realism diagnostics, broader comparisons with temporal causal priors, and learned simulator components.

Finally, the current evaluation does not fully explore all design choices, such as different future-query lengths, nearest-neighbor retrieval baselines, and interpretability of retrieved historical windows. These analyses are important for understanding when and why sample-level attention is most effective.

\end{document}